%% file: main.tex
\documentclass[lettersize,journal]{IEEEtran}
\usepackage{amsmath,amsfonts}
\usepackage{algorithmic}
\usepackage{algorithm}
\usepackage{array}
\usepackage[caption=false,font=normalsize,labelfont=sf,textfont=sf]{subfig}
\usepackage{textcomp}
\usepackage{stfloats}
\usepackage{url}
\usepackage{verbatim}
\usepackage{graphicx}
\usepackage{cite}
\usepackage{multirow}
\usepackage{amssymb}
\usepackage{soul}
\usepackage{colortbl}

\newcommand{\ie}{\textit{i}.\textit{e}.}

\newcommand{\eg}{\textit{e}.\textit{g}.}
\newcommand{\etc}{\textit{etc}}
\newcommand{\figref}[1]{Fig.\ref{#1}}
\newcommand{\tabref}[1]{Tab.\ref{#1}}
\newcommand{\equref}[1]{Eqn.(\ref{#1})}
\newcommand{\secref}[1]{Sec.\ref{#1}}
\newcommand{\etal}{\textit{et al}.}

\usepackage{soul}
\soulregister\cite7
\soulregister\secref7
\soulregister\figref7
\soulregister\tabref7
\soulregister\etal7
\usepackage{colortbl}
\definecolor{hl}{rgb}{.99,.91,.0}

\begin{document}

\title{Simultaneously Recovering Multi-Person Meshes and Multi-View Cameras with Human Semantics}

\author{Buzhen~Huang,
        Jingyi Ju,
        Yuan~Shu,
        and~Yangang~Wang,~\IEEEmembership{Member,~IEEE}
                % <-this % stops a space
\thanks{This work was supported in part by the National Natural Science Foundation of China (No. 62076061) and Natural Science Foundation of Jiangsu Province (No. BK20220127).}
\thanks{Buzhen Huang, Jingyi Ju, Yuan Shu, and Yangang Wang are with the School of Automation, Southeast University, Nanjing, China, 210096. All the authors from Southeast University are affiliated with the Key Laboratory of Measurement and Control of Complex Systems of Engineering, Ministry of Education, Nanjing, China.}% <-this % stops a space
\thanks{Corresponding author: Yangang Wang. E-mail: yangangwang@seu.edu.cn.}
}

\markboth{IEEE TRANSACTIONS ON CIRCUITS AND SYSTEMS FOR VIDEO TECHNOLOGY, VOL. 34, NO. 6, JUNE 2024}%
{Shell \MakeLowercase{\textit{et al.}}: A Sample Article Using IEEEtran.cls for IEEE Journals}


\maketitle

\begin{abstract}
  Dynamic multi-person mesh recovery has broad applications in sports broadcasting, virtual reality, and video games. However, current multi-view frameworks rely on a time-consuming camera calibration procedure. In this work, we focus on multi-person motion capture with uncalibrated cameras, which mainly faces two challenges: one is that inter-person interactions and occlusions introduce inherent ambiguities for both camera calibration and motion capture; the other is that a lack of dense correspondences can be used to constrain sparse camera geometries in a dynamic multi-person scene. Our key idea is to incorporate motion prior knowledge to simultaneously estimate camera parameters and human meshes from noisy human semantics. We first utilize human information from 2D images to initialize intrinsic and extrinsic parameters. Thus, the approach does not rely on any other calibration tools or background features. Then, a pose-geometry consistency is introduced to associate the detected humans from different views. Finally, a latent motion prior is proposed to refine the camera parameters and human motions. Experimental results show that accurate camera parameters and human motions can be obtained through a one-step reconstruction. The code are publicly available at~\url{https://github.com/boycehbz/DMMR}.
\end{abstract}


\input{introduction.tex}

\input{relatedwork.tex}

\input{method.tex}

\input{experiments.tex}

\input{conclusion.tex}

\bibliographystyle{IEEEtran}
\bibliography{human}

% \newpage

 

% \end{IEEEbiography}



% \vspace{-12mm}
\begin{IEEEbiography}[{\includegraphics[width=1in,height=1.25in,clip,keepaspectratio]{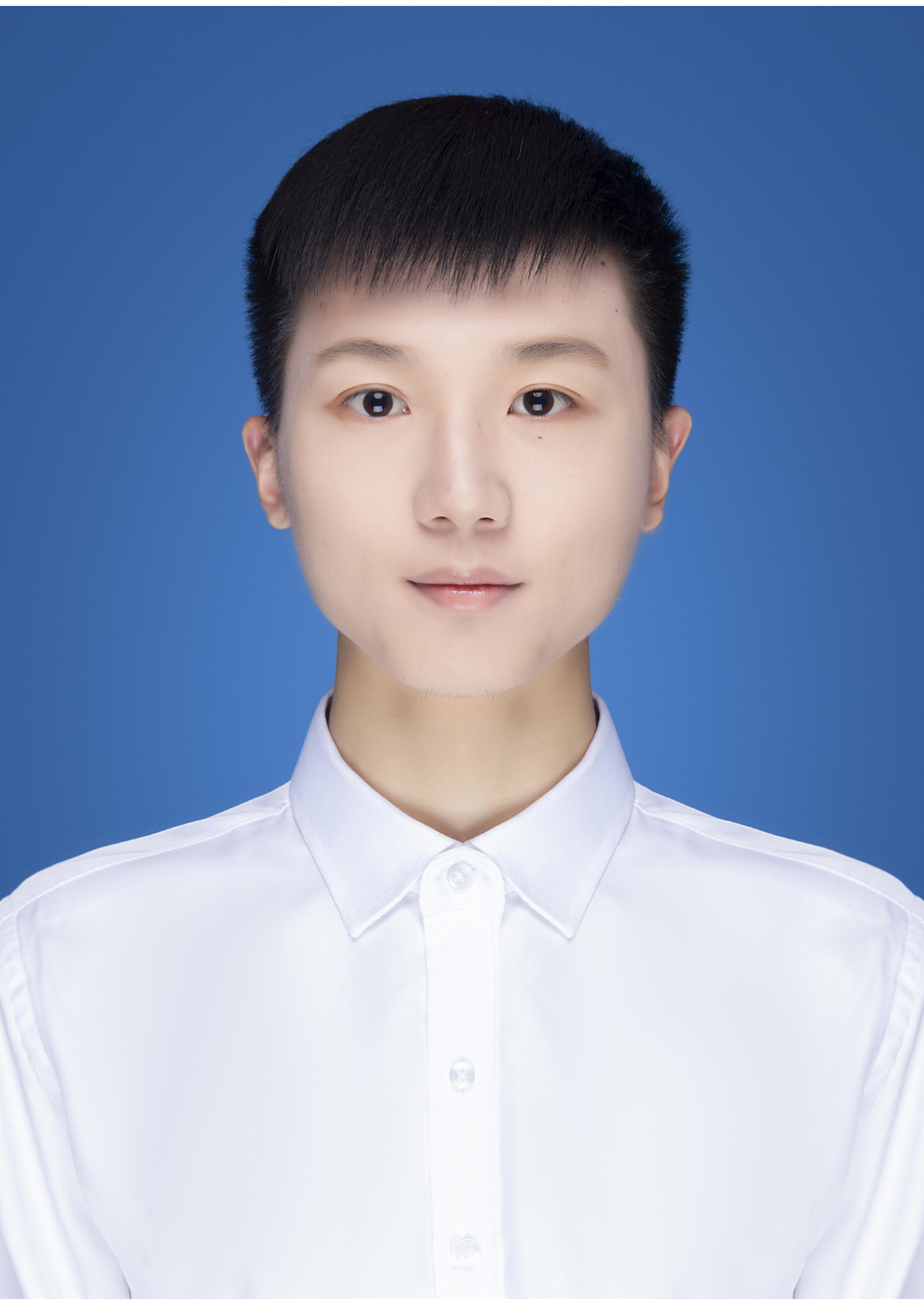}}]{Buzhen Huang} received the bachelor's degree in Automation from Hangzhou Dianzi University, Hangzhou, China, in 2019. He is currently pursuing a Ph.D. degree at Southeast University, Nanjing, China. His research interests include computer vision and computer graphics.
\end{IEEEbiography}
\vspace{-10mm}
\begin{IEEEbiography}[{\includegraphics[width=1in,height=1.25in,clip,keepaspectratio]{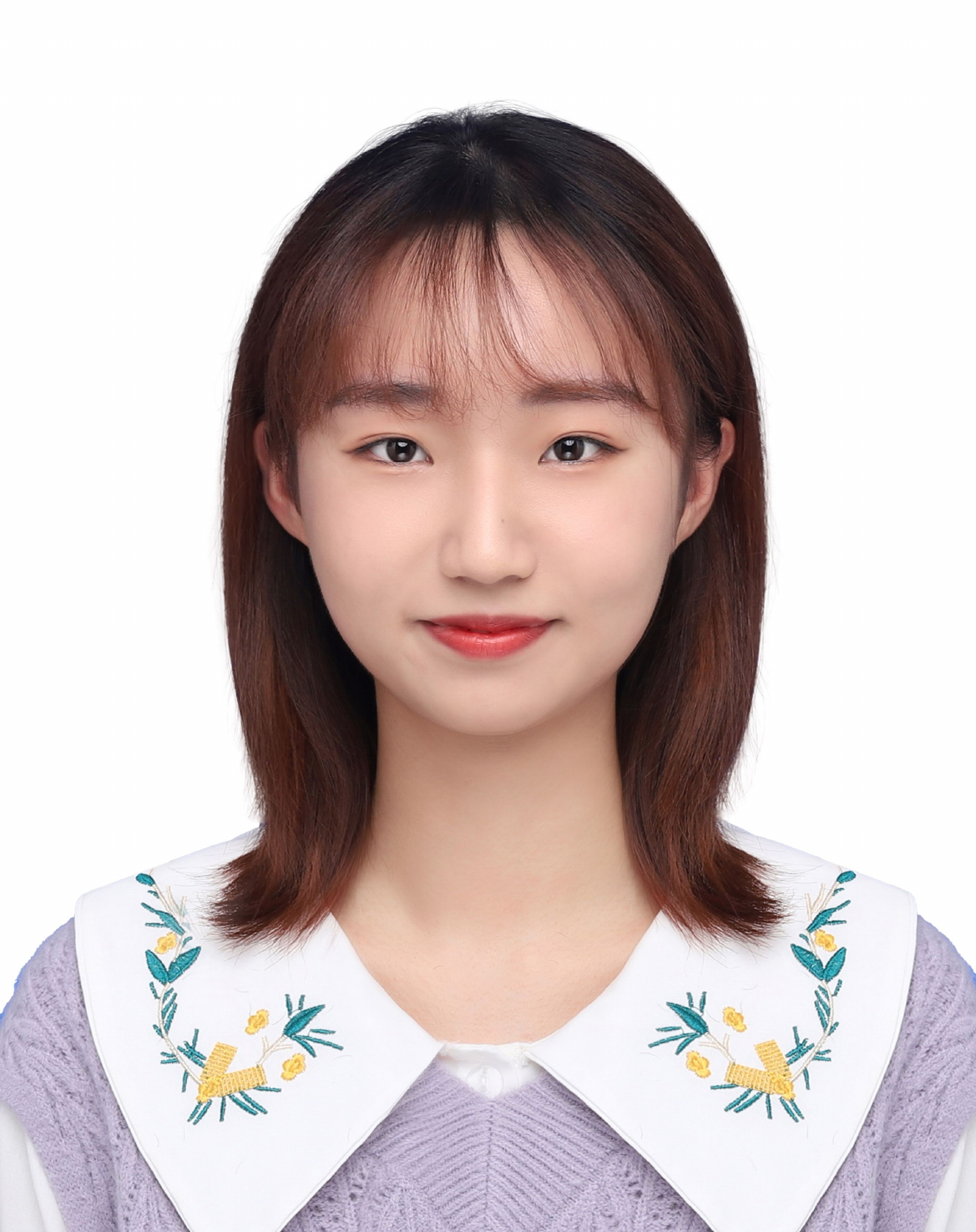}}]{Jingyi Ju} is a graduate student at Southeast University in Vision Cognition Lab, Nanjing, China. Before starting her M.S., she received her bachelor's degree at Nanjing University of Aeronautics and Astronautics, Nanjing, China, in 2022. Her research interests include computer vision and human mesh recovery.
\end{IEEEbiography}
\vspace{-10mm}
\begin{IEEEbiography}[{\includegraphics[width=1in,height=1.25in,clip,keepaspectratio]{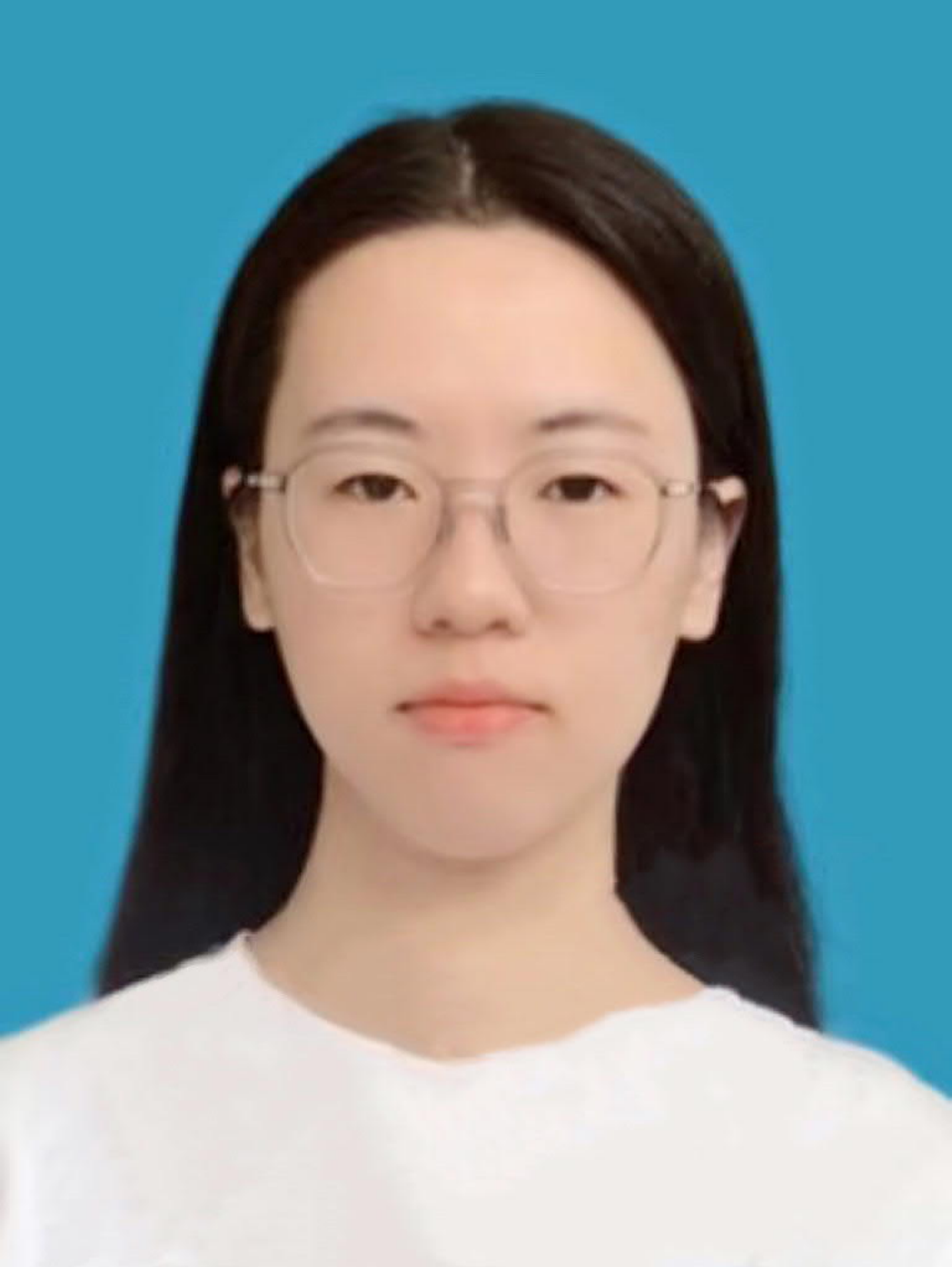}}]{Yuan Shu} received the B.S. degree in Information Science and Engineering from Chien-Shiung Wu college, Southeast University, Nanjing, China in 2023. She is currently working toward the M.S. degree in Information Science and Engineering. Her research interests include signal processing and computer vision.
\end{IEEEbiography}
\vspace{-10mm}
\begin{IEEEbiography}[{\includegraphics[width=1in,height=1.25in,clip,keepaspectratio]{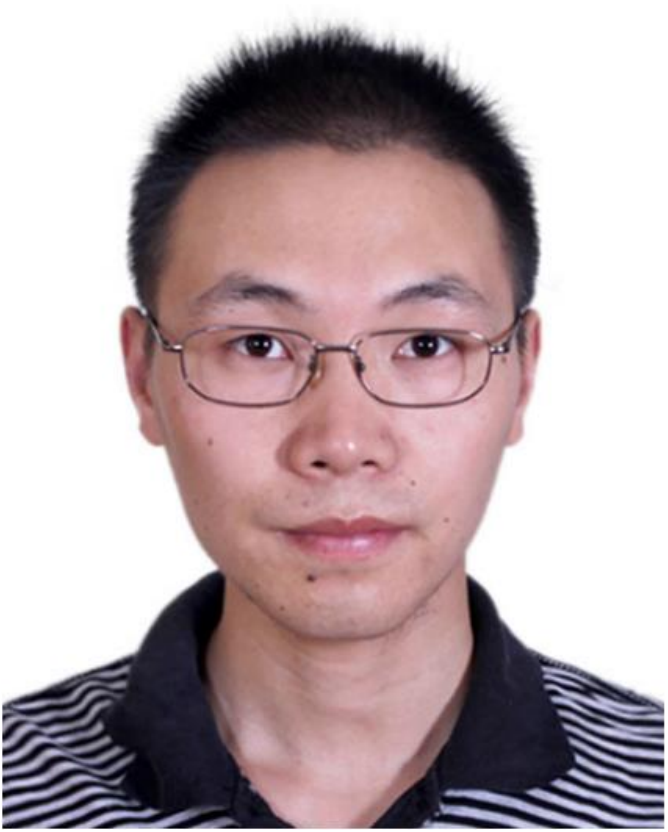}}]{Yangang Wang} received his B.E. degree from Southeast University, Nanjing, China, in 2009 and his Ph.D. degree in control theory and technology from Tsinghua University, Beijing, China, in 2014. He was an associate researcher at Microsoft Research Asia from 2014 to 2017. He is currently an associate professor at Southeast University. His research interests include image processing, computer vision, computer graphics, motion capture and animation.
\end{IEEEbiography}

\vfill

\end{document}

%% file: introduction.tex
\section{Introduction}\label{sec:introduction}

\IEEEPARstart{R}{ecovering} multiple human motions from videos is essential for many applications, such as social behavior understanding, sports broadcasting, virtual reality applications, \etc. Numerous previous works have focused on capturing multi-person motions from multi-view input, employing geometry constraints~\cite{dong2021fast, lin2021multi, zhang20204d, belagiannis20143d, joo2017panoptic, li2011markerless} or optimization-based model fitting~\cite{yang2019spatio,li2020full, li2018shape, liu2013markerless}. While these works have made remarkable advances in multi-person motion capture, they all rely on accurately calibrated cameras~\cite{alexiadis2016integrated} to build view-view and model-view consistency. Few works focus on multi-person motion capture from uncalibrated cameras. Mustafa~\etal~\cite{mustafa2015general} constructed a two-stage framework that first calibrates the camera using the static geometry and then generates 3D human models from dynamic object reconstruction with segmentations. Regression-based pose estimation is also used to find pose pairs for calibration~\cite{ershadi2021uncalibrated}. However, these methods require a large space distance among the target people and can not capture interactive human bodies.

In this paper, we address the problem of simultaneously recovering multiple human body meshes and camera parameters from multi-view videos. There are two primary challenges. The first one is that inter-person interactions and occlusions introduce inherent ambiguities for camera calibration and motion reconstruction. Ambiguous low-level visual features lead to severe low and high frequency noises in detected human semantics~(\eg, 2D pose~\cite{belagiannis20153d}, appearance~\cite{li2020full}), which causes difficulties in establishing view-view and model-view consistency. The other is that the lack of sufficient local image features~(\eg, SIFT~\cite{lowe2004distinctive}) can be used to constrain sparse camera geometries in a dynamic multi-person scene.

To tackle the obstacles, our key-idea is to \textbf{utilize human motion prior knowledge to assist the simultaneous recovery of camera parameters and dynamic human meshes from noisy human semantics.} Specifically, we first detect 2D and 3D poses from the input videos using state-of-the-art pose detectors~\cite{fang2022alphapose,kolotouros2019learning}. The initial intrinsic and extrinsic camera parameters are then estimated from the 2D poses with human cues of upright standing human and cross-view pose similarity. However, due to severe mutual occlusions, the detected poses contain a lot of high-frequency joint jitters, and the identities of each pose cannot be associated via person re-identification~\cite{zhou2019omni,nodehi2021multi} or temporal tracking~\cite{bao2020pose,zou2023snipper}. We therefore propose a pose-geometry consistency based on pose similarity and geometric constraint to further exploit human information to reduce the noises. Compared to previous work~\cite{huang2021dynamic}, the consistency is more robust to less accurate camera parameters and human scale variations. With the consistency, a convex optimization is constructed to associate the human poses from different views and temporal frames.

Once the association is complete, we simultaneously optimize cameras and multi-person motions based on the predicted 2D poses. As the joint optimization is a highly non-convex problem, we introduce a compact latent motion prior to reduce the number of optimization variables. We adopt a variational autoencoder~\cite{kingma2013auto}~(VAE) architecture for our motion prior. Unlike existing VAE-based motion models~\cite{lohit2021recovering, luo20203d, ling2020character}, we use the bidirectional Gate Recurrent Unit (GRU)~\cite{cho2014learning} as the backbone and design a latent space that considers both local kinematics and global dynamics. As a result, our latent prior can be trained on a limited number of short motion clips~\cite{mahmood2019amass} and used to optimize long sequences. Although the motion prior can generate diverse and temporally coherent motions, it is not robust to noise in motion optimization. We found that linearly interpolating the latent code of VPoser~\cite{pavlakos2019expressive} produces consecutive poses. Inspired by this observation, we propose a local linear constraint on motion latent code during model training and optimization. This constraint ensures that the motion prior produces coherent motions from noisy inputs. In addition, to maintain local kinematics, we incorporate a skip-connection between explicit human motion and latent motion code in the model. By using the noisy 2D human semantics as constraints, we can simultaneously recover human motions and camera parameters by optimizing both the latent code and cameras.

\begin{figure*}
    \begin{center}
    \includegraphics[width=0.95\linewidth]{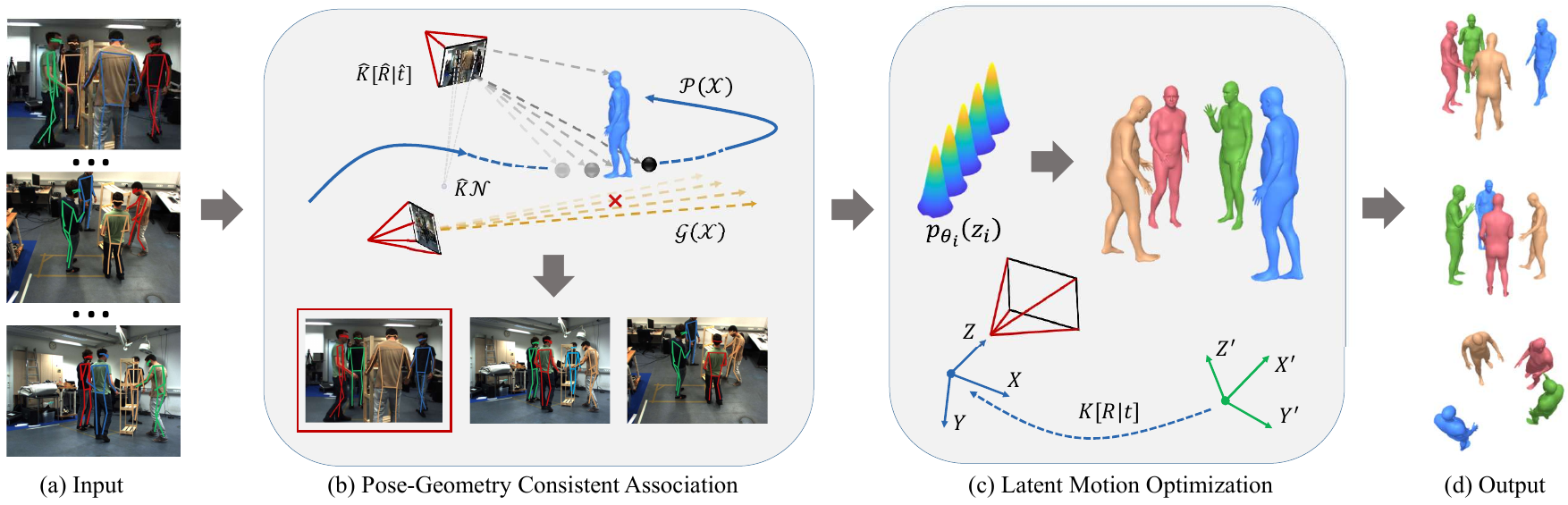}
    \end{center}
    \vspace{-6mm}
    \caption{\textbf{Overview of our method.} Since directly optimizing cameras and human motions from noisy detections~(a) always leads to suboptimal solutions, we first initialize the cameras with human cues. Next, we introduce a pose-geometry consistent association~(b) to establish cross-view and temporal correspondences for the detected human semantics. We further train a latent motion prior for the optimization to obtain accurate camera parameters and coherent human motions from the associated inputs~(d).}
    \label{fig:pipeline}
    \vspace{-6mm}
\end{figure*}

A preliminary version of this paper appeared in \cite{huang2021dynamic}, which introduces a method to recover multi-person motions with extrinsic camera parameters using human semantics. The present work makes the following additional contributions. \textbf{1)} We incorporate a camera intrinsics estimation based on standing human height to construct a complete calibration framework, and thus the method can simultaneously recover human meshes and both intrinsic and extrinsic parameters from multi-view human videos. \textbf{2)} To obtain more stable initial extrinsics, we exploit 3D pose similarity among different views to decompose the extrinsic camera parameters. \textbf{3)} Since the error of the physics-geometry consistency in the previous version~\cite{huang2021dynamic} is proportional to the depth, a person in the distance may be erroneously removed by the denoising procedure. We further improve the consistency with the pose similarity term, which is robust to the human distance and scene size. With the pose consistency, we also avoid the pose tracking required in the previous work~\cite{huang2021dynamic}. \textbf{4)} We also propose a progressive and multi-stage optimization strategy with a pose initialization to improve robustness to the local minima. \textbf{5)} Additional experiments are conducted to prove the effectiveness of each component. We further compare with the latest multi-person mesh recovery approaches to demonstrate the accuracy of our method.

The main contributions of this work are summarized as follows.
\begin{itemize}
    % \vspace{-3mm}
    \item We propose a framework that simultaneously recovers multi-person motions and accurate camera parameters from detected human semantics without the need for any calibration tools.
    \item We propose a calibration-free pose-geometry consistency to establish cross-view and temporal relationships among human semantics.
    \item We propose a compact motion prior trained on short motion clips that can be used for both generating and capturing long motion sequences.
    % \vspace{-3mm}
\end{itemize}

%% file: relatedwork.tex
\section{Related Work}\label{sec:relatedwork}
\noindent\textbf{Multi-view Human pose and shape estimation}. Reconstructing human pose and shape from multi-view inputs has been a long-standing problem. Historically, human segmentation~\cite{liu2016human,zuo2023reconstructing}, depth~\cite{liu2013markerless,alexiadis2016integrated}, appearance~\cite{li2020full} and location~\cite{li2018shape} have been utilized to establish spatio-temporal correspondences for model-based fitting. In contrast, a few works~\cite{belagiannis20143d,belagiannis20153d,lin2021multi,zhang20204d,bridgeman2019multi,joo2017panoptic} first establish view-view correspondences via geometric constraints and then reconstruct through triangulation or optimization. MvPose~\cite{dong2021fast} considers both geometric and appearance constraints simultaneously. However, all these methods rely on accurate camera parameters. Besides, 2D poses and appearance can be easily affected by partial occlusions, which is very common in multi-person interaction sceneries. To recover multiple human meshes from uncalibrated cameras, some works~\cite{mustafa2015general,wang2017outdoor} first calibrate the camera using the static geometry from the background and then generates 3D human models with multiple constraints. Ershadi~\etal~\cite{ershadi2021uncalibrated} realized reconstruction via the similarity of the detected 3D poses from different views. However, these methods require a significant amount of space between the target individuals and cannot capture interactive human bodies.

\noindent\textbf{Extrinsic camera calibration}. Conventional camera calibration methods rely on specific tools~(\eg, checkerboard\cite{zhang2000flexible} and one-dimensional objects\cite{zhang2004camera}), leading to two separate stages for calibration and reconstruction. To address this obstacle, a few works~\cite{zhang2023four,cioppa2021camera,sha2020end,miyata2017extrinsic} estimate camera parameters from the semantics of the scene~(\eg, lines on the basketball court). Other methods obtain the extrinsic camera parameters from the tracked human trajectories~\cite{huang2016camera,tang2019esther}, silhouettes~\cite{ben2016camera}, 2D poses~\cite{garau2019unsupervised}, and human meshes~\cite{garau2020fast} in more general multi-person scenes. Nevertheless, getting accurate human semantics from in-the-wild images itself is a challenging problem. State-of-the-art 2D/3D pose estimation frameworks~\cite{fang2022alphapose,cao2019openpose,kolotouros2019learning} can hardly get accurate 2D/3D keypoints in multi-person scenes, and thus strongly affect the calibration accuracy. To reduce the ambiguities generated by human interactions and occlusions, we propose a pose-geometry consistent association and a robust latent motion prior to remove the noises, realizing multi-person reconstruction and extrinsic camera calibration in an end-to-end way.

\noindent\textbf{Intrinsic camera calibration}. Current intrinsic camera calibration requires a known object~(\eg, checkerboard~\cite{zhang2000flexible} and deltoid grids~\cite{ha2017deltille}) to determine the focal length, which is impracticable to utilize these special tools on the internet images. Therefore, there are some works estimating the intrinsic parameters with the vanishing points~\cite{caprile1990using}. Huang~\etal~\cite{huang2016camera} tracked walking individuals and utilized their trajectories to determine vanishing points or interactions for calibration. However, this method has limitations concerning the types of motion it can handle. In addition, those approaches need to detect parallel lines and solve their intersection points, which would inevitably lead to errors. In contrast to existing works, our method uses the keypoints of upright standing people to initialize intrinsic parameters and refines the results with dense correspondences, avoiding the need for specific long human motions and other restrictions.

\noindent\textbf{Motion prior}. Traditional marker-less motion capture relies on massive views to provide sufficient visual cues~\cite{joo2017panoptic}. Thus, applying strong and compact motion prior~\cite{yang2019spatio} to provide additional knowledge in motion capture has attracted wide attention. Historically, Gaussian Process Latent Variable Model~\cite{yao2011learning} succeeded in modeling human motions since it takes uncertainties into account, but is difficult to make a smooth transition among mixture models. Huang~\etal~\cite{huang2017towards} used a low-dimensional Discrete Cosine Transform basis~\cite{akhter2012bilinear} as the temporal prior to capture human motions. With the development of deep learning, VIBE~\cite{kocabas2020vibe} trains a discriminator to determine the quality of motion, but one-dimensional variables can hardly describe dynamics. Lohit~\etal~\cite{lohit2021recovering} and Luo~\etal~\cite{luo20203d} trained VAEs based on temporal convolutional networks and represent motion with latent code. However, these methods use latent code in a fixed dimension, which is not suitable for dealing with sequences of varying lengths. MotionVAE~\cite{ling2020character} constructs a conditional variational autoencoder to represent motions of the two adjacent frames. Although this structure solves the problem of sequence length variation, it can only model sequence information of the past, which is not suitable for optimizing the whole sequence. In this paper, we propose a motion prior that contains local kinematics and global dynamics of the motion. The structure of the model makes it suitable for large-scale variable-length sequence optimization.

%% file: method.tex
\section{Method}\label{sec:method}

\begin{figure}
    \begin{center}
    \includegraphics[width=0.75\linewidth]{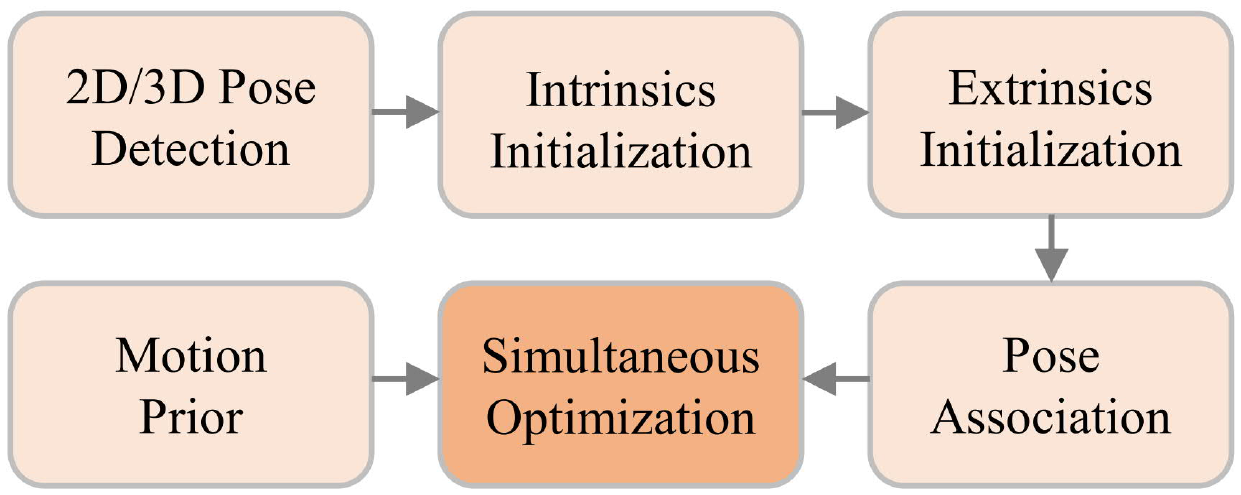}
    \end{center}
    \vspace{-4mm}
    \caption{We show the pipeline of our method and the relationships between different modules. With the aid of a motion prior, our method can simultaneously recover precise camera parameters and human meshes from detected human semantics.}
    \label{fig:pipeline_module}
    \vspace{-6mm}
    \end{figure}

Our goal is to simultaneously recover multiple human motions and camera parameters from multi-view videos. As shown in \figref{fig:pipeline_module}, we first estimate the initial camera intrinsics and extrinsics from detected 2D and 3D poses~(\secref{sec:initialization}). With the initial camera parameters, we then propose a pose-geometry consistency based on pose similarity to identify each person from the detected human semantics~(\secref{sec:association}). We further introduce a latent motion prior~(\secref{sec:motion prior}), which contains human dynamics and kinematics, to assist in the estimation from noisy inputs. Finally, with the trained motion prior, we design an optimization framework with a progressive strategy to simultaneously recover accurate camera parameters and human motions~(\secref{sec:optimization}).

\subsection{Preliminaries}\label{sec:Preliminaries}
\noindent\textbf{Human motion representation}.
We adopt SMPL~\cite{loper2015smpl} to represent human motions, which consists of shape $\beta \in \mathbb{R}^{10}$, pose $\theta \in \mathbb{R}^{72}$ and translation $\mathcal{T} \in \mathbb{R}^{3}$ parameters. To learn a generalized motion prior, we use a 6D representation~\cite{zhou2019continuity} for the body pose and do not consider the global rotation $\mathcal{R} \in \mathbb{R}^{T \times 3}$. Finally, a motion that contains $T$ frames is represented as $\mathcal{X} \in \mathbb{R}^{T \times 138}$.

\noindent\textbf{Camera representation}. We adopt the pinhole camera model and follow previous works~\cite{krahnstoever2005bayesian,liu2011surveillance,fei2021single} by assuming that the principal point of the camera coincides with the image center. Thus, the intrinsic camera parameter can be denoted as:
\begin{equation}
    K = \lbrack f_x, f_y, \frac{w}{2}, \frac{h}{2} \rbrack
\end{equation}
where the $w$ and $h$ are the width and height of the RGB image. The $f_x$ and $f_y$ are focal lengths. We represent the extrinsic camera parameter to be $\mathcal{E} \in \mathbb{R}^{6}$, which contains 3-dimensional rotation and translation, respectively.

\noindent\textbf{2D and 3D pose detection}. We first use an off-the-shelf 2D pose estimation~\cite{fang2022alphapose} to get 2D poses for each person, which will be used to construct the data term of the optimization. In addition, we also estimate the initial 3D pose using~\cite{kolotouros2019learning} for each detected 2D pose. Although the initial 3D pose may be inaccurate, it is sufficient to initialize the camera parameters and construct view-view consistency.

\subsection{Camera Initialization}\label{sec:initialization}
To avoid the use of special calibration tools~(\eg, checkerboard~\cite{zhang2000flexible}), our method only relies on detected human semantics to obtain initial cameras.

\noindent\textbf{Initial intrinsics estimation}. Upright standing humans can provide multiple parallel lines extending from their head keypoints $X_H \in \mathbb{R}^3$ to the midpoint between their ankle keypoints $X_B \in \mathbb{R}^3$. These lines are also perpendicular to the ground plane. To identify upright standing humans among all detections, we leverage the predicted initial 3D poses by evaluating the SMPL parameters of the joints in the spine and legs (\eg, rotations less than a certain threshold). In addition, the corresponding 2D keypoints $x_H$ and $x_B$ can be obtained through the 2D pose detection, which can be utilized to calculate vanishing points~\cite{caprile1990using} for intrinsic parameter estimation. The vanishing points will be the intersection points of all the lines extending from $x_H$ to $x_B$ in the 2D image plane. However, due to the inherent inaccuracy of the initial 2D and 3D poses, the results may not be reliable. To address this issue, we employ a RANSAC algorithm for further filtering. In this process, we first select two parallel lines to calculate a vanishing point in the 2D image plane. To determine the correctness of a line from another individual, we compute the center point between the head and middle ankle keypoints in the 2D image. Subsequently, we establish a new line connecting the vanishing point and the center point. If the angle $\gamma$ between the new and original lines exceeds a predefined threshold, we remove the individual from the list of upright standing human candidates. This procedure is iteratively executed for all samples, following the principles of the RANSAC algorithm.

With the estimated vanishing points, we follow~\cite{fei2021single} to use these keypoints to estimate initial intrinsics. In contrast to~\cite{fei2021single}, which assumes a constant human height, we can obtain the height $h$ from the detected 3D pose, making it applicable to more generalized scenarios. Without loss of generality, we first shift the 2D keypoints with the camera principal point and denote the 2D kyepoints in homogeneous representation with $\hat{x}_H$ and $\hat{x}_B$, and the intrinsic matrix takes the form $\hat{K} = \operatorname{diag}\left(f_{x}, f_{y}, 1\right)$. Thus, we have the projection equations: $\lambda_{H,i} \hat{x}_{H,i} = \hat{K}X_{H,i}$ and $\lambda_{B,i} \hat{x}_{B,i} = \hat{K}X_{B,i}$, where $\lambda$ represents the unknown depth of the keypoint. Another constraint can be derived from the human height:
\begin{equation}\label{eq::difference}
    \lambda_{H,i} \hat{x}_{H,i} - \lambda_{B,i} \hat{x}_{B,i} = \textit{h}_i \cdot \hat{K}\mathcal{N},
\end{equation}
where $\mathcal{N}$ is the ground normal, and $\textit{h}$ is the human height from the detected 3D pose. The unknown depth $\lambda$ can be eliminated by multiplying $\hat{x}_{H,i} \times \hat{x}_{B,i}$:
\begin{equation}
    \left(\hat{x}_{H,i} \times \hat{x}_{B,i}\right) \hat{K}\mathcal{N} = 0.
\end{equation}
With more than two people, we can solve the function by the least square method, and $\hat{K}\mathcal{N}$ is the vertical vanishing point, which can be denoted as $v \triangleq \hat{K}\mathcal{N}$. Since the ankle centers $x_B$ of different standing humans can be spanned to the ground plane, which is orthogonal to the ground plane normal. We then use this constraint to decompose the intrinsic parameters.
\begin{equation}
    v^\textup{T} \left(\lambda_{B,i}\hat{x}_{B,i} - \lambda_{B,j}\hat{x}_{B,j} \right) \hat{K}^{-\textup{T}}\hat{K}^{-1} = 0 ,
\end{equation}
where $ \hat{K}^{-\top} \hat{K}^{-1} \triangleq \operatorname{diag}\left(1 / f_{x}^{2}, 1 / f_{y}^{2}, 1\right)$ can be obtained with more than three people. It should be noted that when the captured people are less than 3, we can still use the same person from different frames to calibrate the camera. The method with only the detected keypoints can significantly simplify the current intrinsic camera calibration system, and we then use the parameters as the initial values for the subsequent procedures. 

\noindent\textbf{Initial extrinsics estimation}. Since the triangulation in the previous work~\cite{huang2021dynamic} requires human identities and is not robust to estimate acceptable initial values. In this work, we improve the extrinsics estimation with the pose similarity. We first rigidly align the 3D poses from different views to build the correspondences with aligned joint errors. The rotation transformations among the paired poses from different views can be regarded as the inverse transformations of the cameras. By assigning the extrinsic parameters of the first camera to be an identity matrix, we can obtain the rotation for the other cameras with the transformations. Then, the translation of each camera can be regressed by the 2D pose and the corresponding 3D pose with the initial intrinsics~\cite{huang2022pose2uv}. Although the parameters may not be accurate, it is adequate for the initial values for the association~(\secref{sec:association}) and joint optimization~(\secref{sec:optimization}).

\begin{figure}
    \begin{center}
    \includegraphics[width=1\linewidth]{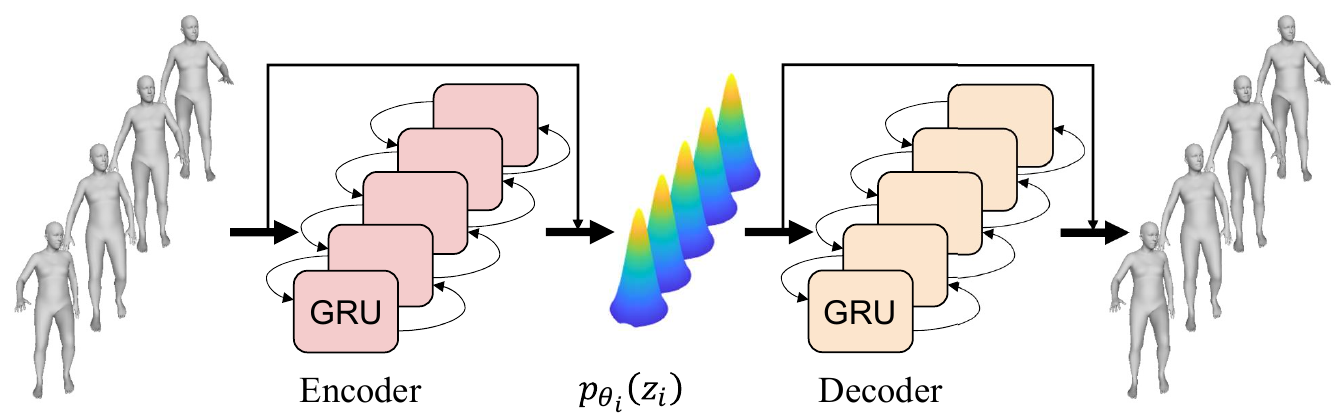}
    \end{center}
    \vspace{-6mm}
    \caption{The motion prior is a symmetrical encoder-decoder network, which compactly models human dynamics and kinematics. The prior can be trained on short clips and used to fit long sequences.}
    \label{fig:motionprior}
    \vspace{-6mm}
    \end{figure}

\subsection{Pose-geometry Consistent Association}\label{sec:association}
We then propose a pose-geometry consistent association to build model-view and view-view correspondences for the detected 2D semantics. The previous work~\cite{huang2021dynamic} requires the users to manually identify each human in different views and then uses a pose tracking~\cite{zhou2019omni,zhou2018deep} with a denoising framework to associate the humans in different frames. The method loses a lot of valid poses that just have the wrong identity. In addition, the error of the physics-geometry consistency in the denoising framework is proportional to the distance from the person to the camera, which is ineffective in a large-scale scene. Therefore, we replace the physics term~\cite{huang2021dynamic} with a pose similarity term and propose a pose-geometry consistent association to avoid the limitations.

Given 2D poses with corresponding 3D poses, we use the similarity of 3D poses to associate the 2D poses. For a detected 3D pose $X_i$ from the view $i$, we first transform the pose to the world camera system with the initial extrinsic parameters $\mathcal{E}_i$. Thus, the temporal pose similarity can be defined as:
\begin{equation}\label{eq:temproalpose}
    \mathcal{L}^{i}_{t} = || X^{t-1} - \hat{X}_{i} ||,
\end{equation}
where $X^{t-1}$ is the 3D pose in the previous frame and $\hat{X}_{i}$ is the transformed $X_i$. $\mathcal{L}^{i}_{t}$ measures the similarity between $X^{t-1}$ and $\hat{X}_{i}$, which is employed to associate the most similar pose among all detections in adjacent frames. In addition, the spatial pose similarity between view $i$ and view $j$ is:
\begin{equation}
    \mathcal{L}^{i,j}_{s} = || \hat{X}_{i} - \hat{X}_{j} ||.
\end{equation}
We average the paired 3D poses to obtain $X^{t}$ to calculate the temporal pose similarity in the next frame~(\equref{eq:temproalpose}).

However, with only the above constraints, the people that have similar poses may be erroneously associated. Thus, we further utilize a set of optical rays, which come from the optical center of the camera and pass through corresponding 2D joint coordinates, to construct an additional geometric term to constrain the spatial position. For view $i$, the ray in the pl\"ucker coordinates is represented as $(n_{i}, l_{i})$. We enforce the rays from view $i$ and view $j$ to be coplanar precisely:
\begin{equation}
    \mathcal{L}_{g}^{i,j} = n_{i}^T l_{j} + n_{j}^T l_{i}.
\end{equation}

We combine these constraints as the pose-geometry consistency to associate the detected poses from different views. The pose cost and geometric cost of different views are represented in matrices $\mathcal{P}$ and $\mathcal{G}$.

\begin{equation}\label{eq:association}
    \left\{
        \begin{aligned}
        \mathcal{P}_{i,j} &= \mathcal{L}_{t}^{i} + \mathcal{L}_{t}^{j} + \mathcal{L}_{s}^{i,j} \\
        \mathcal{G}_{i,j} &= \mathcal{L}_{g}^{i,j}
        % y & = & \sin(t) \\
        % z & = & \frac xy
        \end{aligned}
    \right.,
\end{equation}
where $\mathcal{P}_{i,j}$ and $\mathcal{G}_{i,j}$ are pose cost and geometric cost of view $i$ and view $j$. We use a positive semidefinite matrix $\mathcal{M} \in \{0,1\}^{v \times v} $~\cite{huang2013consistent} to represent the correctness of correspondences among different views. Our goal is to solve $\mathcal{M}$, which minimizes the pose-geometry consistency cost:
\begin{equation}
    \mathop{\arg\min}_{\mathcal{M}} f(\mathcal{M})=-c_{g}\langle\mathcal{G}, \mathcal{M}\rangle-c_{p}\langle\mathcal{P}, \mathcal{M}\rangle,
\end{equation}
where $c_{g}$, $c_{p}$ are 0.7 and 0.3 in our experiment. $\langle \rangle$ denotes the hadamard product. A semi-positive definite matrix $\mathcal{M}$ can ensure the correspondences between different views to be correct, thus reducing the influence of the noises. During the iteration, the $\mathcal{M}$ is a real matrix whose element values are in the range of 0 - 1. The alternating direction method of multipliers~\cite{boyd2011distributed} is adopted to solve this problem. We choose the result with the highest number of corresponding views in $\mathcal{M}$ as the final output, which we subsequently utilize for joint optimization. Any results lacking corresponding views are considered incorrect detections. In conclusion, we employ the estimated $\mathcal{M}$ to extract accurate detections.

In the first frame, we do not apply the~\equref{eq:temproalpose} in the~\equref{eq:association} and then average the paired 3D poses to obtain $X_0$. Thus, the view-view correspondences for each human can be automatically built, which avoids the manually identifying in the previous work~\cite{huang2021dynamic}. After the association, the filtered 2D poses will be used in \equref{equ:data} to find optimal motions.

\subsection{Latent Motion Prior}\label{sec:motion prior}
With the initial cameras and associated 2D poses, we can optimize cameras and SMPL models by a non-linear fitting. However, simultaneous optimization of multi-person motions and camera parameters from noisy 2D poses is a highly non-convex problem and is likely to get stuck in the local minima. To address this challenge, we design a compact VAE-based latent motion prior to obtain accurate and temporal coherent motions. The prior has three strengths. 1) It contains compact dynamics and kinematics to reduce computational complexity. 2) It can be trained on short motion clips and applied to long sequence fitting. 3) The latent local linear constraint ensures robustness to noisy input. The details are described as follows.

\noindent\textbf{Model architecture}. 
Our network is based on VAE~\cite{kingma2013auto}, which shows great power in modeling motions~\cite{ling2020character, luo20203d}. As shown in~\figref{fig:motionprior}, the encoder consists of a bidirectional GRU, a mean and variance encoding network with a skip-connection. The decoder has a symmetric network structure. Different from previous work~\cite{ling2020character}, the bidirectional GRU ensures that the prior is able to see all the information from the entire sequence and that the latent code can represent global dynamics. However, the latent prior encoded only by features extracted from GRU is difficult to reconstruct accurate local fine-grained poses when used for large-scale sequence optimization. Thus, we construct a skip-connection for the encoder and decoder, respectively, allowing the latent prior to accurately capture the refined kinematic poses and the global correlation between them. Besides, we design the latent code $\mathbf{z} \in \mathbb{R}^{T \times 32}$ whose frame length $T$ corresponds to the input sequence. Thus, our prior can be trained on a limited amount of short motion clips~\cite{mahmood2019amass} and be applied to long sequence fitting.

\noindent\textbf{Training}. 
In the training phase, a motion $\mathcal{X}$ is fed into the encoder to generate mean $\mu\left(\mathcal{X}\right)$ and variance $\sigma\left(\mathcal{X}\right)$. The sampled latent code $\mathbf{z} \sim q_{\phi}\left(\mathbf{z} \mid \mu\left(\mathcal{X}\right), \sigma\left(\mathcal{X}\right)\right)$ is then decoded to get the reconstructed motion $\mathbf{\hat{\mathcal{X}}}$. The reparameterization trick~\cite{kingma2013auto} is adopted to achieve gradient backpropagation. We train the network through maximizing the Evidence Lower Bound (ELBO):

\begin{equation}
    \begin{array}{l}
    \log p_{\theta}\left(\mathcal{X}\right) \geq \mathbb{E}_{q_{\phi}}\left[\log p_{\theta}\left(\mathcal{X}\mid\mathbf{z}\right)\right] \\
    -D_{\mathrm{KL}}\left(q_{\phi}\left(\mathbf{z}\mid\mathcal{X}\right) \| p_{\theta}\left(\mathbf{z}\right)\right).
    \end{array}
\end{equation}

The specific loss function is:
\begin{equation}
    \mathcal{L}_{vae} = \mathcal{L}_{\mathrm{6d}} + \mathcal{L}_{\mathrm{v}} + \mathcal{L}_{\mathrm{kl}}+ \mathcal{L}_{\mathrm{linear}},
\end{equation}
where $\mathcal{L}_{6d}$ and $\mathcal{L}_{v}$ are:
\begin{equation}
    \mathcal{L}_{6d}=\sum_{t=1}^{T}\left\|\mathcal{X}_{t}-\mathbf{\hat{\mathcal{X}}}_{t}\right\|^{2},
\end{equation}

\begin{equation}
    \mathcal{L}_{v}=\sum_{t=1}^{T}\left\|\mathbf{\mathcal{V}}_{t}-\mathbf{\hat{\mathcal{V}}}_{t}\right\|^{2},
\end{equation}

where $\mathcal{V}_{t}$ is the deformed SMPL vertices of frame $t$. $\mathcal{L}_{v}$ guarantees that the prior learns high fidelity local details.
\begin{equation}
    \mathcal{L}_{kl}=K L(q(\mathbf{z}\mid\mathcal{X}) \| \mathcal{D}(0, I)),
\end{equation}

Although applying the above constraints can produce diverse and temporal coherent motions, it is not robust to noisy 2D poses. The jitter and drift of 2D poses and identity error will result in an unsmooth motion. Inspired by the interpolation of VPoser~\cite{pavlakos2019expressive}, we add a local linear constraint to enforce a smooth transition on latent code:
\begin{equation}
    \mathcal{L}_{linear}=\sum_{t=1}^{T-1} \|z_{t+1}-2z_{t}+z_{t-1}\|.\label{equ:linear}
\end{equation}

When the motion prior is applied in long sequence fitting, the parameters of the decoder are fixed. The latent code is decoded to get the motion $\hat{\mathcal{X}} \in \mathbb{R}^{T \times 138}$.

\subsection{Joint Optimization of Motions and Cameras}\label{sec:optimization}

 \noindent\textbf{Optimization variables}. Unlike traditional structure-from-motion~(SFM), which lacks structural constraints among 3D points and is vulnerable to noisy input. We directly optimize the motion prior, so that the entire motions are under inherent kinematic and dynamic constraints. The optimization variables of $V$ views videos that contain $N$ people are $\{\left(\beta, \mathbf{z}, \mathcal{R}, \mathcal{T}\right)_{1: N}, \left(\mathcal{E}, K\right)_{1:V}\}$.

\noindent\textbf{Objective}. 
We formulate the objective function as follows:
\begin{equation}
    \mathop{\arg\min}_{\left(\beta, \mathbf{z}, \mathcal{R}, \mathcal{T}\right)_{1: N}, \left(\mathcal{E}, K\right)_{1:V}} \mathcal{L}=\mathcal{L}_{data}+\mathcal{L}_{prior}+\mathcal{L}_{pen},
\end{equation}

where the data term is:
\begin{equation}
    \mathcal{L}_{data} =\sum_{v=1}^{V} \sum_{n=1}^{N} \sigma_{v}^{n} \rho\left(\Pi_{\mathcal{E}_{v}, K_v}\left(\mathrm{J}^{n}\right)-\mathbf{p}_{v}^{n}\right)\label{equ:data}
\end{equation}

where $\rho$ is the robust Geman-McClure function~\cite{geman1987statistical}. $\mathbf{p}$, $\sigma$ are the filtered 2D poses and its corresponding confidence. $\mathrm{J}$ is the skeleton joint position generated by model parameters.

Besides, the regularization term is:
\begin{equation}
    \mathcal{L}_{prior} = \sum_{n=1}^{N} \left\|\mathbf{z}_{n}\right\|^{2} + \sum_{n=1}^{N} \left\|\beta_{n}\right\|^{2} + \sum_{n=1}^{N} \mathcal{L}_{linear}.
\end{equation}

$\mathcal{L}_{linear}$ is the same as~\equref{equ:linear}. We further apply a collision term based on differentiable Signed Distance Field~(SDF)~\cite{jiang2020coherent} to prevent artifacts generated from multi-person interactions.

\begin{equation}
    \mathcal{L}_{pen}=\sum_{j=1}^{N} \sum_{i=1, i \neq j}^{N} \sum_{vt \in \mathcal{V}_{j}} -\min (\operatorname{SDF}_{i}(vt), 0),
\end{equation}
where $\operatorname{SDF}(vt)$ is the distance from sampled vertex $vt$ to the human mesh surface.

\noindent\textbf{Optimization}. The multivariate optimization is a challenging problem even with the proposed compact motion prior. To make the optimization to be tractable, we further propose several optimization strategies. 1) We use the initial 3D poses to initialize the SMPL parameters before the optimization. 2) We partition the entire sequence into multiple batches, with each batch comprising $F$ frames. We optimize $F$ frames for the first time and progressively add more batches in the optimization objective until all frames are optimized. 3) We employ a strategy inspired by SMPLify~\cite{2016Keep}, dividing the optimization into 3 substages. The camera parameters, global human positions, and body poses are optimized in the different substages. The above strategies make the optimization to be more robust and accurate.

\begin{figure*}
    \begin{center}
    \includegraphics[width=1.0\linewidth]{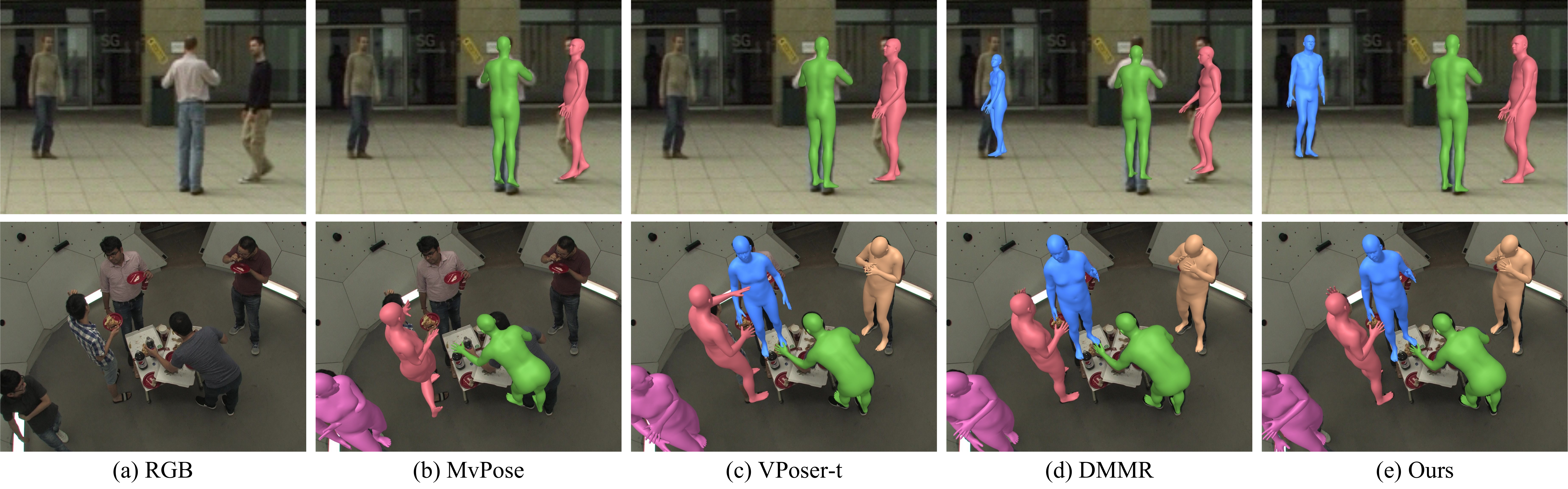}
    \end{center}
    \vspace{-6mm}
    \caption{Qualitative comparison with multi-view methods on Campus~(Row 1) and Panoptic~(Row 2) datasets. Campus captures humans in a large scene~(we zoom in for better visualization). DMMR cannot reconstruct humans in the distance, and MvPose~\cite{dong2021fast} also fails on these cases due to the mismatched 2D pose and the lack of prior knowledge.}
    \label{fig:qulitative_comparison}
\vspace{-5mm}
\end{figure*}

%% file: experiments.tex
\section{Experiments}\label{sec:experiment}
In this section, we conduct several evaluations to demonstrate the effectiveness of our method. The comparisons in \secref{sec:mocap experiment} demonstrate that our method is capable of recovering multiple human bodies from videos and achieves a state-of-the-art performance. Then, we prove that the accurate camera parameters can be obtained from joint optimization in \secref{sec:camera experiment}. Finally, several ablations in \secref{sec:ablation} are conducted to evaluate key components.

\subsection{Datasets}\label{sec:datasets}
\textbf{Campus and Shelf}~\cite{belagiannis20143d} The Campus and Shelf datasets contain more than 3 characters with partial occlusions and across-view ambiguities. We follow the same evaluation protocol as in previous works~\cite{dong2019fast} and compute the PCP (percentage of correctly estimated parts) scores to measure the accuracy of 3D pose estimation.

\textbf{Panoptic}~\cite{joo2015panoptic} This dataset is captured in a studio with 480 VGA cameras and 31 HD cameras, which contains multiple people engaging in social activities. We conduct qualitative and quantitative experiments on \textit{160906\_pizza1} sequence with HD cameras.

\textbf{MHHI}~\cite{liu2011markerless} is a multi-person dataset that contains complex and extreme poses as well as fast motion. The Fight sequence is publicly available and captured in a marker-based manner. For a fair comparison, quantitative experiments are conducted on this sequence.

\textbf{OcMotion}~\cite{huang2022object} is a 3D dataset that contains object-occluded humans. We evaluate our method on this dataset to reveal the superiority of our approach on occluded monocular scenarios. We use the sequence \textit{0013, 0015, 0017, 0019} for quantitative evaluation.

\textbf{Human3.6M}~\cite{ionescu2013human3} is a large-scale, single human dataset captured in a controlled scene, which consists of 11 subjects with 4 views. It provides accurate 3D joint positions and camera parameters. We follow~\cite{huang2017towards, li20213d} to use S9 and S11 for evaluation.

\textbf{3DHP}~\cite{mehta2017monocular} is captured using a multi-view system. The standard testset contains 1 view under indoor and outdoor scenes. We use the testset to evaluate our method in the single-view setting.

\textbf{AMASS}~\cite{mahmood2019amass} is a large collection of 15 motion capture datasets with a unified SMPL representation. The dataset is used for motion prior training.

\textbf{vPTZ}~\cite{possegger2012unsupervised} is a dataset captured with three (outdoor) to four (indoor) static Axis P1347 cameras, as well as one spherical Point Grey Ladybug3 camera. It captures surveillance scenarios. We use it to evaluate the estimated cameras.

\textbf{3DMPB}~\cite{huang2022pose2uv} contains multi-view images with close-range viewpoints. We adopt this dataset to evaluate our method in more complex camera settings.

\textbf{MvMHAT}~\cite{gan2021self} is a benchmark for the multi-view multi-human association and tracking task. We use its testing set with static cameras to evaluate pose-geometry consistent association.

\subsection{Implementation Details}\label{sec:Implementation}
Due to the limited human motion data, we use data augmentation to enhance the generalization performance of the model when training the motion prior. The strategy mainly includes 1) Upsampling and downsampling. We upsample or downsample the origin sequences to generate motions in different frame rates. 2) Reverse sampling. We sample the sequence from the end frame to the start frame to generate a new sequence. 3) Flip sampling. Since the human body is symmetrical, we generate new motions by following the kinematic tree of the human model to mirror the motion across the left and right. We train the prior using short motion clips with 16 frames.

Although the motion prior is compact, jointly optimizing large-scale multi-person motion sequences and camera parameters is still a highly non-convex problem. To reduce the solution space, we use the initial camera parameters to initialize the global positions and rotations of human models in each frame. The coarse 3D skeleton joint positions are first triangulated. Then we rigidly align the models to the estimated 3D joints. The rotations and translations of the aligned models are used as initial values for the joint optimization. Our optimization is implemented in PyTorch ~\cite{paszke2017automatic} using L-BFGS~\cite{nocedal2006numerical} optimizer. Due to the pose initialization, the joint optimization is accelerated compared to DMMR~\cite{huang2021dynamic}. On a desktop with an Intel(R) Core(TM) i9-9900K CPU and a GPU of NVIDIA GeForce RTX 2080Ti, 30s 5-view RGB videos with 4 people take about 5.5 min to fit, in which the pose-geometry consistent association takes about 40s and joint optimization spends 4.8min.

\begin{figure*}
    \begin{center}
    \includegraphics[width=1.0\linewidth]{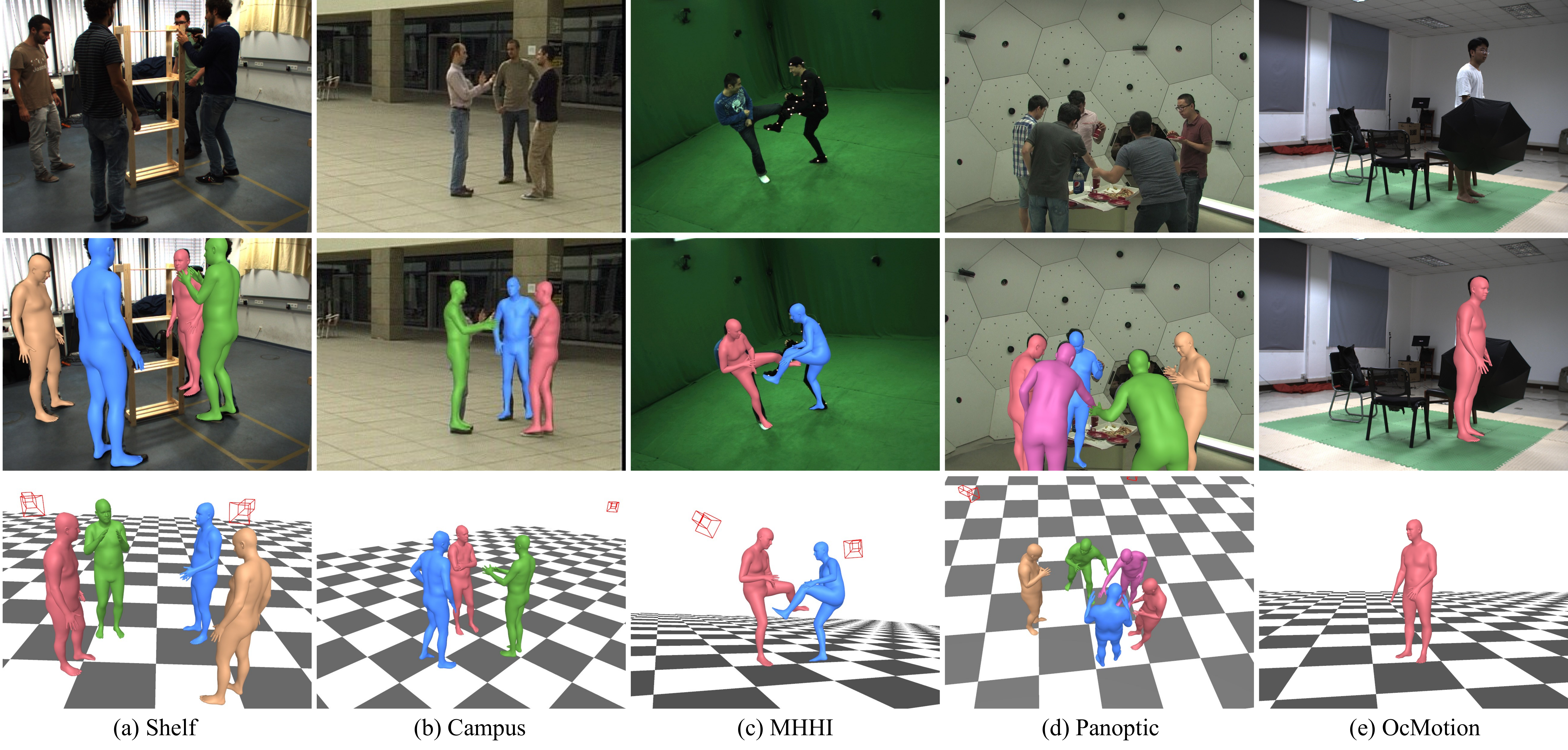}
    \end{center}
    \vspace{-6mm}
    \caption{We show more results on different datasets. Our method can estimate accurate cameras and motions in a one-step reconstruction.}
    \label{fig:qulitative_results}
\vspace{-5mm}
\end{figure*}

\begin{table}
\begin{center}
    \caption{\textbf{Comparison with baseline methods that estimate multi-person 3D poses.} The numbers represent the percentage of correctly estimated parts~(PCP). "VPoser-t" is a combination of VPoser~\cite{pavlakos2019expressive}.}
    \label{tab:campus_shelf}
    \vspace{-5mm}
    \resizebox{1.0\linewidth}{!}{
        \begin{tabular}{l|c c c|c c c}
        \noalign{\hrule height 1.5pt}
        \begin{tabular}[l]{l}\multirow{2}{*}{Method}\end{tabular}
        &\multicolumn{3}{c|}{\textit{Campus}} &\multicolumn{3}{c}{\textit{Shelf}} \\
        &A1   &A2  & A3 &A1 &A2  &A3 \\
        \noalign{\hrule height 1pt}
        Belagiannis~\etal~\cite{belagiannis20143d} &82.0 &72.4 &73.7 &66.1 &65.0 &83.2 \\
        Belagiannis~\etal~\cite{belagiannis20153d} &93.5 &75.7 &85.4 &75.3 &69.7 &87.6 \\
        Bridgeman~\etal~\cite{bridgeman2019multi}  &91.8 &92.7 &93.2 &99.7 &92.8 &97.7 \\
        Dong~\etal~\cite{dong2021fast} & 97.6 &93.3  &98.0 & 98.9 & 94.1 &97.8  \\
        Chen~\etal~\cite{chen2020cross} &97.1 &\textbf{94.1} &98.6 &99.6 &93.2 &97.5 \\
        Zhang~\etal~\cite{zhang20204d} &-- &-- &-- &99.0 &96.2 &97.6 \\
        Chu~\etal~\cite{chu2021part} &\textbf{98.4} &93.8 &98.3 &99.1 &95.4 &97.6 \\
        \hline\hline
        Zhang~\etal~\cite{zhang2021lightweight} &-- &-- &-- &99.5 &\textbf{97.0} &97.8 \\
        Zhang~\etal~\cite{zhang2021direct} &98.2 &\textbf{94.1} &97.4 &99.3 &95.1 &97.8 \\
        Dong~\etal~\cite{dong2021shape} &-- &-- &-- &99.1 &93.5 &\textbf{98.1} \\
        VPoser-t~\cite{pavlakos2019expressive} &97.3 &93.5 &98.4 &\textbf{99.8} &94.1 &97.5 \\
        DMMR~\cite{huang2021dynamic} &97.6 &93.7 &\textbf{98.7} &\textbf{99.8} &96.5 &97.6 \\
        \textbf{Ours}&98.1 &93.6 &\textbf{98.7} &\textbf{99.8} &96.5 &\textbf{98.1} \\
        \noalign{\hrule height 1.5pt}
        \end{tabular}
    }
\end{center}
\vspace{-5mm}
\end{table}

\subsection{Human Motion Capture from RGB Videos}\label{sec:mocap experiment}
We first conducted qualitative and quantitative comparisons on Campus and Shelf datasets to demonstrate the accuracy of captured human motions. Several baseline methods that regress 3D poses are compared. Belagiannis~\etal introduced the concept of 3D pictorial structure for multi-person 3D pose estimation, applied to both multi-view images~\cite{belagiannis20143d} and videos~\cite{belagiannis20153d}. Other recent works~\cite{bridgeman2019multi, dong2021fast, chen2020cross, zhang20204d, chu2021part} are based on calibrated cameras. They directly estimate joint positions and thus have a smaller deviation in joint definition than mesh-based methods, which inherently can achieve more accurate results. The quantitative results in \tabref{tab:campus_shelf} demonstrate that our method can also outperform them in some cases on Campus and Shelf datasets in terms of PCP. In \figref{fig:qulitative_comparison}~(Row 1), the Campus dataset captures a large scene, and the humans are in the distance. We found that DMMR does not produce satisfactory results for individuals positioned far from the camera since the error of the filtering is proportional to the depth change. While our method is more robust with the pose-geometry consistent association. We further compared to the latest mesh-based methods~\cite{zhang2021lightweight,zhang2021direct,dong2021shape,huang2021dynamic} in \tabref{tab:campus_shelf}. Some of them also use numerical optimization to achieve multi-person motion capture~\cite{zhang2021lightweight,zhang2021direct,huang2021dynamic}. Due to the pose-geometry association and motion prior, our method can still achieve state-of-the-art in most cases.

\begin{table}
    \begin{center}
        \caption{\textbf{Quantitative comparison with multi-person mesh recovery methods on MHHI dataset.} The numbers represent the mean distance between markers and their paired 3D vertices in \textit{mm}.}
        \label{tab:MHHI}
        \vspace{-5mm}
        \resizebox{1\linewidth}{!}{
            \begin{tabular}{l|c c c c c}
            \noalign{\hrule height 1.5pt}
            \begin{tabular}[l]{l}\multirow{1}{*}{Method}\end{tabular}
            
            &1 view  &2 views & 4 views&8 views &12 views  \\
            \noalign{\hrule height 1pt}

            Liu~\etal~\cite{liu2013markerless} & - &- &- & - & 51.67  \\
            Li~\etal~\cite{li2018shape} &1549.88&242.27&58.42 &48.57 &43.30  \\
            Li~\etal~\cite{li2020full} &- & 63.93&37.88 & 32.73&30.35  \\
            VPoser-t~\cite{pavlakos2019expressive} &158.33 &60.02 &38.46 &32.11 &31.48  \\
            DMMR~\cite{huang2021dynamic} &140.96 &58.04 &37.86 &30.92 &29.83  \\
            \textbf{Ours} &\textbf{123.64} &\textbf{53.41}  &\textbf{36.83}  &\textbf{30.72}  &\textbf{29.80}   \\

            \noalign{\hrule height 1.5pt}
            \end{tabular}
        }
\end{center}
\vspace{-5mm}
\end{table}

\begin{table}
    \begin{center}
        \vspace{-5mm}
        \caption{\textbf{Comparison on single-person datasets.} "MPJPE" and "PA-MPJPE" are measured in $mm$. "Accel" represents acceleration error in $mm/s^2$. Our method produces more accurate and temporally coherent results. "$^*$" denotes learning-based regression.}\label{tab:3doh}
        \vspace{-5mm}
        \resizebox{1.0\linewidth}{!}{
            \begin{tabular}{l|c c c|c c c}
            \noalign{\hrule height 1.5pt}
            \begin{tabular}[l]{l}\multirow{2}{*}{Method}\end{tabular}
            
            &\multicolumn{3}{c|}{\textit{Human3.6M}} &\multicolumn{3}{c}{\textit{3DHP}} \\
            &PA-MPJPE   &MPJPE  &Accel &PA-MPJPE &MPJPE  &Accel\\
            \noalign{\hrule height 1pt}
            $^*$Li~\etal~\cite{li20213d}              &43.8 &64.8 &--   &65.1 &97.6  &--  \\
            $^*$Liang~\etal~\cite{liang2019shape}     &45.1 &79.9 &--   &--   &62.0  &-- \\
            Huang~\etal~\cite{huang2017towards}       &47.1 &58.2 &--   &--   &--    &-- \\
            VPoser-t                                  &34.7 &53.5 &10.2 &73.5 &103.6 &105.4 \\
            w/o local linear                          &32.5 &44.2 &7.1  &64.7 &99.1 &43.7 \\
            Ours                                      &30.3 &43.3 &4.6  &62.4 &90.2 &32.3 \\
            \noalign{\hrule height 1.5pt}
            \end{tabular}
        }
\end{center}
\vspace{-8mm}
\end{table}

\begin{figure}
    \begin{center}
    \includegraphics[width=1.0\linewidth]{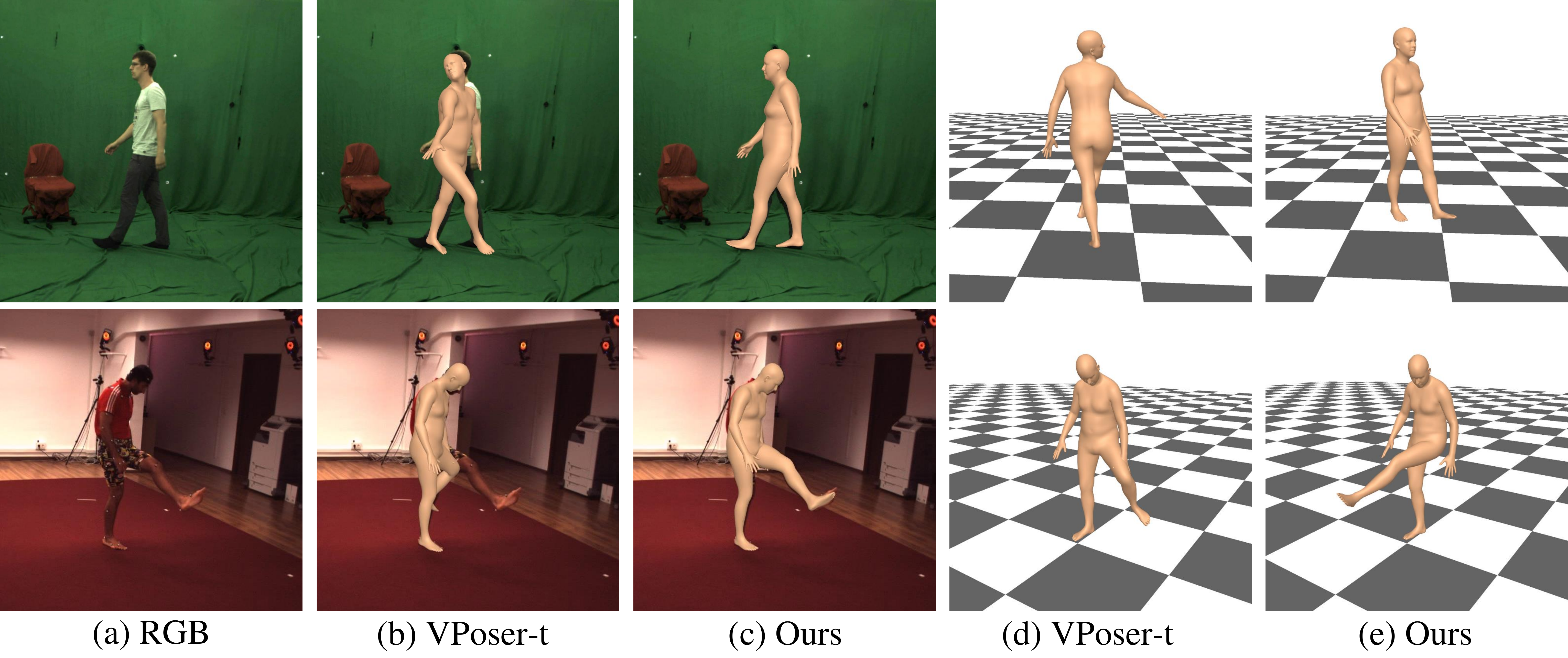}
    \end{center}
    \vspace{-5mm}
    \caption{Comparsion with VPoser-t on single-view cases.}
    \label{fig:quli_compare}
\vspace{-5mm}
\end{figure}

We then evaluated our method on MHHI dataset. Only a few works~\cite{liu2013markerless,li2018shape,li2020full} have successfully reconstructed closely interacting humans from multi-view input. However, all of these works rely on accurately calibrated camera parameters. We conducted quantitative comparisons with these methods in \tabref{tab:MHHI}. In the single-view case, since the motion prior provides additional prior knowledge, our method generates far more accurate results than \cite{li2018shape}. In addition, the proposed approach achieves competitive results with fewer views. Besides, we further compared with DMMR in this dataset. Since we use the initial poses for initialization, our method significantly improves the performance and robustness on sparse cameras.

Li~\etal~\cite{li20213d} and Liang~\etal~\cite{liang2019shape} trained neural networks to regress SMPL parameters from multi-view images. Huang~\etal~\cite{huang2017towards} proposed an optimization-based method for fitting the human model to multi-view 2D keypoints. We compared our method with these baselines on single-person datasets. As shown in \tabref{tab:3doh}, we reported the mean per joint position error~(MPJPE), the MPJPE after rigid alignment of the prediction with ground truth using Procrustes Analysis~(PA-MPJPE) to evaluate the accuracy of estimated skeleton joints. Furthermore, we used the acceleration error~(Accel), which is calculated as the difference in acceleration between the ground-truth and predicted 3D joints, to describe the quality of the predicted motion. The results in~\tabref{tab:3doh} demonstrate that our method achieves state-of-the-art. Besides, with the local linear constraint, the acceleration error decreases by 2.5 on Human3.6M dataset, proving that it produces more coherent motions.

To further demonstrate the effectiveness of the proposed method on single-view cases, we show the qualitative results on 3DHP and Human3.6M in \figref{fig:quli_compare}. Due to the ambiguity of symmetrical skeleton joints in the side view, VPoser-t cannot use temporal information to penalize incoherent results. With the local kinematics and global dynamics in the motion prior, our method can produce more accurate results in these situations.

\subsection{Camera Calibration Evaluation}\label{sec:camera experiment}
% \vspace{-2mm}
We then conducted experiments to evaluate our camera calibration. We first compared the accuracy of the intrinsic camera calibration for initial values with other methods on vPTZ dataset~\cite{possegger2012unsupervised}. As mentioned in~\secref{sec:Preliminaries}, the image origin coincides with the camera's principal point, and we do not focus on distortion calibration. So that we follow previous works~\cite{fei2021single,krahnstoever2005bayesian} to use the normalized deviation of the estimated focal length from the ground-truth focal length, \ie, $\frac{|\hat{f}-f|}{f} \times 100 \%$, to evaluate our intrinsics estimation, where $\hat{f}$ and $f$ are estimated and true focal lengths. The quantitative results are shown in \tabref{tab:init intrinsic calibration}, which reveal a significant improvement compared with previous works. Liu~\etal~\cite{liu2011surveillance} used noisy foreground masks as input. However, they relied on prior knowledge of the relative distribution of human heights to achieve robust camera calibration. Other works~\cite{brouwers2016automatic, tang2019esther, fei2021single, krahnstoever2005bayesian} assume that human height remains constant. In contrast, we estimate the human height through the human mesh, which is estimated by~\cite{kolotouros2019learning}, so we can get a more precise human height for each person and achieve more accurate intrinsics prediction.

\begin{table}
    \begin{center}
        \caption{\textbf{Evaluation of the estimated intrinsic parameters on vPTZ dataset.} The numbers are focal length errors in \%.}
        \label{tab:init intrinsic calibration}
        \vspace{-4mm}
        \resizebox{1\linewidth}{!}{
            \begin{tabular}{l|c |c |c }
                \noalign{\hrule height 1.5pt}
                \begin{tabular}[l]{l}\multirow{1}{*}{Method}\end{tabular} &\begin{tabular}[1]{c}Outdoor \\ \textit{cam131}\end{tabular} &\begin{tabular}[1]{c}Outdoor \\ \textit{cam132}\end{tabular} &\begin{tabular}[1]{c}Indoor \\ \textit{cam132}\end{tabular}\\
                \noalign{\hrule height 1pt}
                Liu~\etal~\cite{liu2011surveillance}                &1  &29     &-       \\
                Brouwers~\etal~\cite{brouwers2016automatic}         &-  &15     &24.92   \\
                ESTHER~\cite{tang2019esther}                        &-  &10.14  &12.07   \\
                Fei~\etal~\cite{fei2021single}                      &4.7&0.35   &10.74   \\
                Krahnstoever~\etal~\cite{krahnstoever2005bayesian}  &-  &9.47   &-       \\
                Initial ours                                             &\textbf{0.96} &\textbf{0.26} &\textbf{3.90}  \\
                \noalign{\hrule height 1.5pt}
                \end{tabular}
            }
\end{center}
\vspace{-6mm}
\end{table}

\begin{figure}
    \begin{center}
    \includegraphics[width=1.0\linewidth]{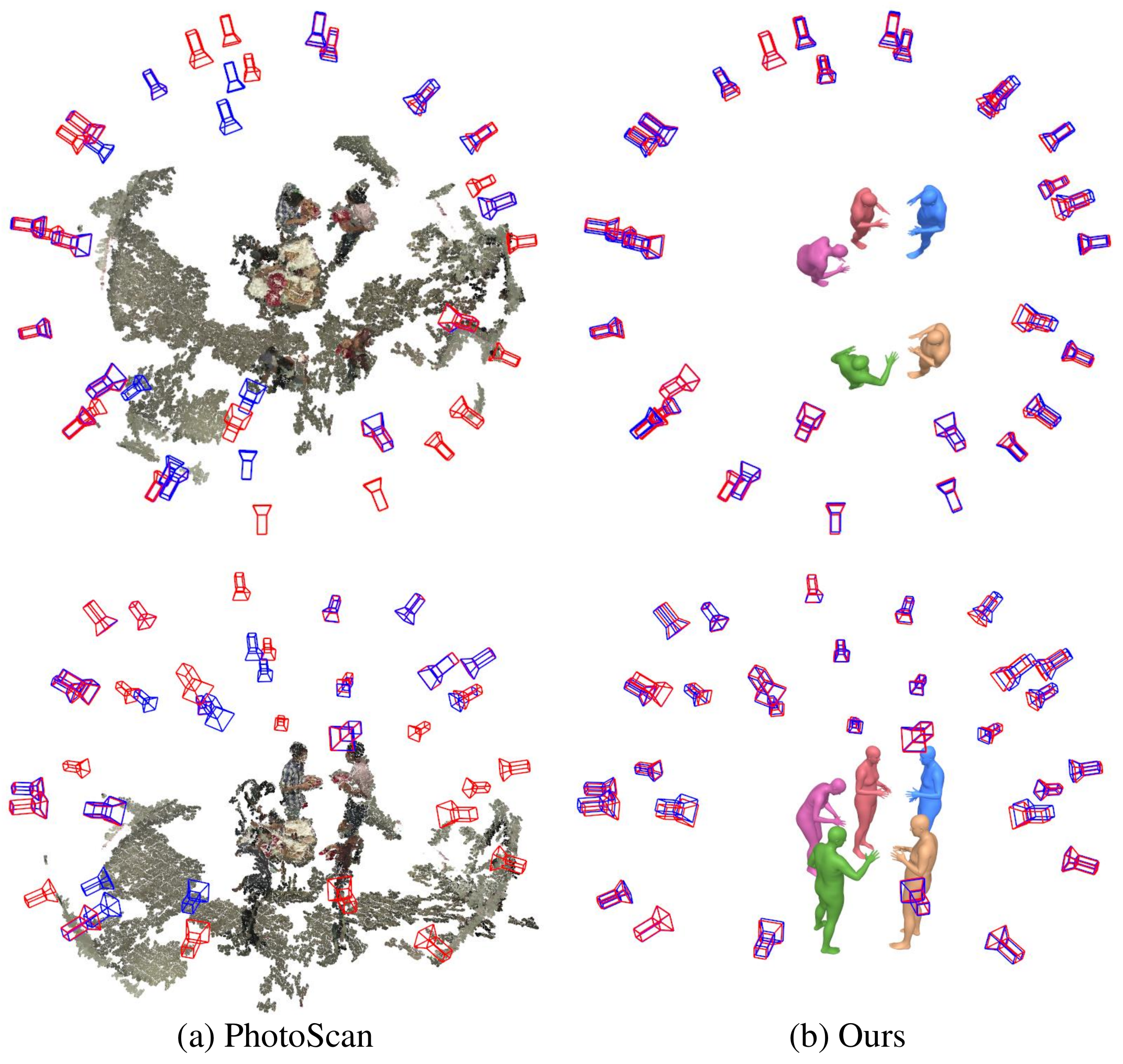}
    \end{center}
    \vspace{-6mm}
    \caption{We conducted a comparison with PhotoScan on Panoptic using 31-view input. Our method accurately estimates all camera parameters from noisy human semantics, whereas PhotoScan only obtains a subset of cameras.}
    \label{fig:qulitative_cam}
    \vspace{-6mm}
    \end{figure}

\begin{table*}
    \begin{center}
        \caption{\textbf{Evaluation of estimated camera parameters.} The "$focal$" value represents the focal length error in percentage. "\textit{Pos.}" and "\textit{Ang.}" denote the position error and angle error between the predicted cameras and the ground-truth camera parameters, measured in millimeters (\textit{mm}) and degrees (\textit{deg}), respectively. "\textit{Reproj.}" is re-projection error in pixels. The "Initial" values correspond to the coarse camera parameters estimated in \secref{sec:initialization}. Additionally, we employ the notations "\textit{+ext.}" and "\textit{+int.}" to indicate simultaneous optimization of extrinsic and intrinsic camera parameters. We also compare our method with a separate estimation approach referred to as "Separate ours," which initially refines the initial camera parameters using noisy 2D poses and subsequently reconstructs 3D body meshes with the refined cameras.}
        \vspace{-5mm}
        \label{tab:camera Calibration}
        \resizebox{1.0\linewidth}{!}{
            \begin{tabular}{l|c c c c|c c c c}
            \noalign{\hrule height 1.5pt}
            \begin{tabular}[l]{l}\multirow{2}{*}{Method}\end{tabular} &\multicolumn{4}{c|}{Panoptic Dataset} &\multicolumn{4}{c}{Shelf Dataset} \\
            
            % &\textit{Positional error} &\textit{Angular error} &\textit{Re-projection error}\\
            &\textit{\begin{tabular}[1]{c}$focal$ \\ (\%)\end{tabular}} &\textit{\begin{tabular}[1]{c}Pos.\\(mm)\end{tabular}} &\textit{\begin{tabular}[1]{c} Ang. \\ (deg)\end{tabular}} &\textit{\begin{tabular}[1]{c} Reproj. \\ (pixel)\end{tabular}} &\textit{\begin{tabular}[1]{c}$focal$ \\ (\%)\end{tabular}} &\textit{\begin{tabular}[1]{c}Pos.\\(mm)\end{tabular}} &\textit{\begin{tabular}[1]{c} Ang. \\ (deg)\end{tabular}} &\textit{\begin{tabular}[1]{c} Reproj. \\ (pixel)\end{tabular}}\\
            % &initial   &final & initial &final & final & ground-truth\\
            \noalign{\hrule height 1pt}
            Ground-truth &- &- &- &22.51  &- &- &- &18.50    \\
            PhotoScan        &5.04 &505.02 &35.29 &188.18 &- &- &- &-   \\
            Initial DMMR~\cite{huang2021dynamic}   &- &3358.51 &44.30 &637.21 &- &1532.42 &26.86 &79.34 \\
            Initial ours           &5.50 &2603.47  &28.77  &266.96  &8.15 &605.88  &9.72  &44.94  \\
            Separate ours      &5.50 &945.64  &2.81  &28.41  &7.74 &55.14  &2.10  &18.73  \\
            VPoser-t~\cite{pavlakos2019expressive}  \textit{+ext.}   &- &118.88 &0.64 &22.76 &- &34.30 &0.59 &18.83   \\

            DMMR~\cite{huang2021dynamic} \textit{+ext.}           &- &101.25 &0.59 &22.69 &- &23.18 &0.52 &18.70  \\
            \textbf{Ours} \textit{+ext.}       &- &116.29  &0.67  &\textbf{22.56}  &- &23.22 &0.50 &18.63   \\
            VPoser-t~\cite{pavlakos2019expressive}  \textit{+int +ext.}   &5.50 &834.31  &1.68  &23.18  &7.14 &44.26 &2.11 &14.35   \\
            % \hline
            DMMR~\cite{huang2021dynamic}  \textit{+int +ext.}   &5.50 &723.01  &1.47  &23.10  &7.11 &32.89 &1.78 &14.57   \\
            \textbf{Ours} \textit{+int +ext.}  &5.47 &635.88  &1.34  &22.89  &6.14 &26.45 &1.32 &\textbf{11.33}   \\
            \noalign{\hrule height 1.5pt}
            \end{tabular}
        }
\end{center}
\vspace{-6mm}
\end{table*}

\begin{figure*}
    \begin{center}
    \includegraphics[width=1.0\linewidth]{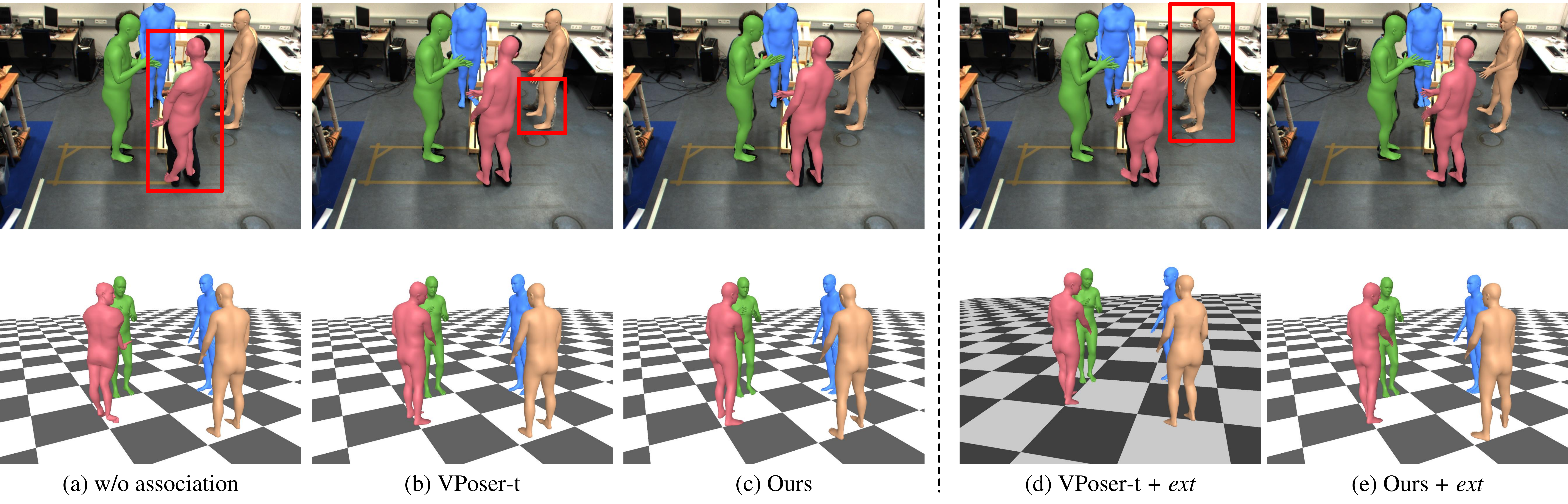}
    \end{center}
    \vspace{-7mm}
    \caption{\textbf{Ablation study on pose-geometry consistent association and motion prior.} Without the association, the optimization cannot obtain accurate motion due to the influence of noises. Due to the absence of motion dynamics, VPoser-t is hard to estimate plausible cameras and motions when the cameras are not provided.} 
    \label{fig:ablation}
\vspace{-6mm}
\end{figure*}

\begin{table*}
    \begin{center}
        \caption{\textbf{Ablation study on pose-geometry consistent association and motion prior.} "Mean" and "Std" represent the mean distance and corresponding standard deviation between the vertices and markers.}
        \label{tab:ablation}
        \vspace{-6mm}
        \resizebox{1.0\linewidth}{!}{
            \begin{tabular}{l|c c|c c c|c c c|c c c}
            \noalign{\hrule height 1.5pt}
            \begin{tabular}[l]{l}\multirow{2}{*}{Method}\end{tabular}
            
            &\multicolumn{2}{c|}{\textit{MHHI}} &\multicolumn{3}{c|}{\textit{Campus}} &\multicolumn{3}{c|}{\textit{Shelf}} &\multicolumn{3}{c}{\textit{OcMotion}}\\
            &Mean$\downarrow$   &Std$\downarrow$  &A1$\uparrow$ &A2$\uparrow$ &A3$\uparrow$   &A1$\uparrow$ &A2$\uparrow$ &A3$\uparrow$  &MPJPE$\downarrow$ &PA-MPJPE$\downarrow$ &PVE$\downarrow$ \\ 
            \noalign{\hrule height 1pt}
            VPoser-t~\cite{pavlakos2019expressive} &31.48 &11.54 &97.3 &93.5 &98.4 &\textbf{99.8} &94.1 &97.5  &99.6 &60.9 &102.8\\
            DMMR w/o filtering~\cite{huang2021dynamic} &32.31 &12.17 &96.7 &93.0 &96.8 &92.4 &89.8 &91.6  &- &- &-\\
            DMMR w/o local linear~\cite{huang2021dynamic} &30.25 &11.07 &97.0 &93.5 &98.4 &\textbf{99.8} &95.4 &97.3  &99.3 &61.8 &101.0\\
            DMMR~\cite{huang2021dynamic}  &29.83 &\textbf{9.87} &97.6 &\textbf{93.7} &\textbf{98.7} &\textbf{99.8} &\textbf{96.5} &97.6  &94.2 &57.3 &99.6\\
            Ours &\textbf{29.80}  &9.89  &\textbf{98.1} &93.6 &\textbf{98.7} &\textbf{99.8}  &\textbf{96.5}  &\textbf{98.1}   &\textbf{88.2} &\textbf{53.4} &\textbf{95.3}\\
            \hline\hline
            VPoser-t~\cite{pavlakos2019expressive} \textit{+ext.} &43.72 &19.57 &89.4 &83.2 &84.1 &97.4 &89.7 &89.7  &- &- &-\\
            DMMR w/o filtering~\cite{huang2021dynamic} \textit{+ext.} &49.34 &24.37 &86.3 &82.1 &84.0 &91.5 &86.7 &88.6 &- &- &-\\
            DMMR w/o local linear~\cite{huang2021dynamic} \textit{+ext.}&35.25 &17.07 &92.4 &82.3 &83.4 &97.5 &90.4 &93.3 &- &- &-\\
            DMMR~\cite{huang2021dynamic} \textit{+ext.} &34.44 &10.57 &92.3 &84.6 &85.1 &\textbf{98.4} &91.5 &94.4 &- &- &-\\
            Ours \textit{+ext.} &\textbf{33.26}  &\textbf{10.52}  &\textbf{93.1} &\textbf{86.7} &\textbf{90.1}    &\textbf{98.4}  &\textbf{92.1}  &\textbf{94.6}  &- &- &-\\
            \hline\hline
            Separate ours &63.61  &23.01  &87.2 &83.2 &82.6 &93.8  &82.6  &83.5   &- &- &-\\
            VPoser-t~\cite{pavlakos2019expressive} \textit{+int +ext.} &56.21  &20.11  &88.2 &83.0 &83.4 &96.7  &86.6  &86.5   &- &- &-\\
            DMMR~\cite{huang2021dynamic} \textit{+int +ext.} &39.45  &13.23  &88.1 &84.3 &86.3 &96.9  &87.4  &88.6  &- &- &-\\
            Ours \textit{+int +ext.} &\textbf{37.78}  &\textbf{12.31}   &\textbf{90.7} &\textbf{86.8} &\textbf{89.8}    &\textbf{97.8}  &\textbf{90.1}  &\textbf{93.6}  &- &- &-\\

            \noalign{\hrule height 1.5pt}
            \end{tabular}
        }
\end{center}
\vspace{-8mm}
\end{table*}

We also qualitatively and quantitatively evaluate the estimated camera parameters on Panoptic and Shelf dataset. Since a rigid transformation exists between the predicted camera parameters and the ground-truth, we follow \cite{cioppa2021camera} to apply rigid alignment to the estimated cameras. We first compared with PhotoScan~\footnote{https://www.agisoft.com/}, a commercial software that reconstructs 3D point clouds and cameras. As shown in \tabref{tab:camera Calibration}, PhotoScan fails to work for sparse inputs~(Shelf dataset) since it relies on the dense correspondences between each view. We evaluate the results with position, angle, and re-projection errors. Under relatively massive views, our method outperforms PhotoScan in all metrics. \figref{fig:qulitative_cam} shows the results on Panoptic dataset with 31 views. The cameras in red and blue colors are the ground-truth and the predictions, respectively. PhotoScan only captures part of the cameras with low accuracy. On the contrary, our method successfully estimates all the cameras with complete human meshes. In addition, the SMPL skeleton and motion prior can provide inherent geometric constraints (\ie, bone length and symmetry) and motion prior knowledge for the 3D points. These constraints are crucial for camera calibration using noisy 2D poses. To further demonstrate the effectiveness of our simultaneous optimization, we compare it with a separate estimation method. This method initially refines the initial camera parameters using noisy 2D poses and then reconstructs 3D body meshes with the refined cameras. We found that our method is more robust to noisy inputs, and thus can obtain more accurate cameras on Panoptic and Shelf datasets. Compared to the previous version, we use pose information from different views to estimate the initial extrinsics, which is more robust and accurate than triangulation. With the joint optimization, the final results gain significant improvement. Our method achieves better performance both from massive and sparse inputs with the pose-geometry association and the motion prior. The re-projection error is the most important metric in calibration. On Shelf dataset, the ground-truth cameras are not very accurate. We found that our method with intrinsics estimation even outperforms the ground-truth.

\begin{table}
    \begin{center}
        \caption{\textbf{Ablation Study of Motion Prior Components.} "w/ basic GRU" indicates the use of a single GRU instead of the bidirectional GRU. "w/o skip-connection" and "w/o local linear" refer to the motion prior without the skip-connection and local linear constraint.}
        \label{tab:ablation_motionprior}
        \vspace{-3mm}
        \resizebox{0.7\linewidth}{!}{
            \begin{tabular}{l|c c}
            \noalign{\hrule height 1.5pt}
            \begin{tabular}[l]{l}\multirow{2}{*}{Method}\end{tabular}
            
            &\multicolumn{2}{c}{\textit{MHHI}} \\
            &Mean$\downarrow$   &Std$\downarrow$   \\ 
            \noalign{\hrule height 1pt}
            VPoser-t~\cite{pavlakos2019expressive} &31.48 &11.54 \\
            w/ basic GRU &31.96 &9.93 \\
            w/o skip-connection &32.34 &10.34 \\
            w/o local linear &30.10 &11.14 \\
            Motion Prior &\textbf{29.80}  &\textbf{9.89}  \\
            
            \noalign{\hrule height 1.5pt}
            \end{tabular}
        }
\end{center}
\vspace{-6mm}
\end{table}

\begin{table}
    \begin{center}
        \caption{We assess the precision of multi-view subject association~(AIDP)~\cite{gan2021self} using the MvMHAT dataset to demonstrate the effectiveness of our pose-geometry consistent association approach. The association method used for MvPose also relies on our initial camera parameters. When compared to DMMR, we observe that the spatial pose similarity cost plays a crucial role in cross-view association.}
        \label{tab:association}
        \vspace{-5mm}
        \resizebox{1\linewidth}{!}{
            \begin{tabular}{c|c c c c}
            \noalign{\hrule height 1.5pt}
            \begin{tabular}[l]{l}\multirow{1}{*}{Method}\end{tabular}
            
            &MvPose~\cite{dong2021fast}  &DMMR~\cite{huang2021dynamic} &MvMHAT~\cite{gan2021self} &Ours   \\
            \noalign{\hrule height 1pt}

            AIDP &26.3 &39.4  &53.0  &\textbf{55.1}   \\

            \noalign{\hrule height 1.5pt}
            \end{tabular}
        }
\end{center}
\vspace{-8mm}
\end{table}

% \vspace{-3mm}
\subsection{Ablation Study}\label{sec:ablation}
\begin{figure}
    \begin{center}
    \includegraphics[width=1.0\linewidth]{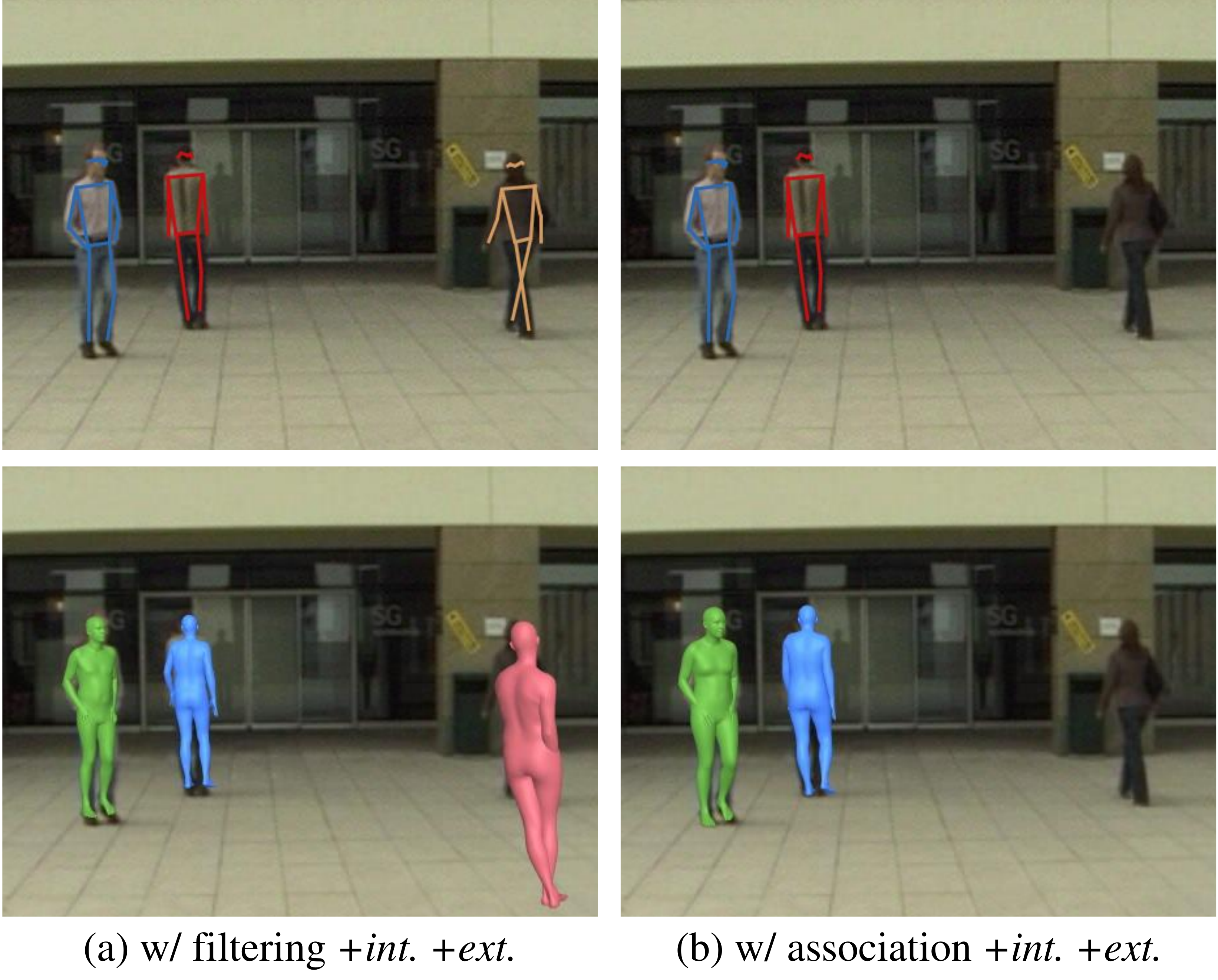}
    \end{center}
    \vspace{-6mm}
    \caption{The filtering in DMMR~\cite{huang2021dynamic} is not robust for humans standing far away from the camera, while the association is generalized to depth and inaccurate cameras.}
    \label{fig:association}
    \vspace{-3mm}
\end{figure}

% \vspace{-2mm}
\noindent\textbf{Pose-geometry consistent association}. We conducted ablation on the pose-geometry consistent association to reveal its significance in associating noisy human semantics. In \tabref{tab:ablation}, we first remove the filtering based on the physical-geometry consistency proposed in DMMR. Thus, the input 2D poses contain many noisy detections. The results on Shelf dataset show that the noises severely affect the motion capture performance for sparse camera inputs. DMMR applies the filtering and achieves better results. However, it relies on optical lines and is not robust to depth variations and inaccurate cameras. \figref{fig:association} shows the filtering cannot get correct results for the people in the distance. When the initial intrinsic parameters are inaccurate, the performance of the filtering is strongly affected due to the incorrect optical lines. On the contrary, our association is robust since the rigid pose alignment is independent of the focal length. In addition, the filtering also discards some detections, which have the correct 2D pose and the wrong identity. To fully exploit the 2D detections, we replace the filtering with the association. The poses with the same identity are selected from the detections, and thus all valid poses can be used to constrain human motions. We further evaluate our pose-geometry consistent association using the standard cross-view pose matching benchmark, the MvMHAT dataset~\cite{gan2021self}. Previous cross-view pose association methods have primarily been employed in multi-view human tracking~(\eg, MvMHAT~\cite{gan2021self}), relying on human appearance cues for association. In contrast, our approach leverages 3D information to enhance the association process. Benefited from advanced 3D pose estimators, our association method with the spatial pose similarity cost is more robust to different camera views. The results, presented in \tabref{tab:association}, demonstrate that our method significantly outperforms DMMR and also surpasses MvMHAT in terms of AIDP.

\begin{figure}
    \begin{center}
    \includegraphics[width=1.0\linewidth]{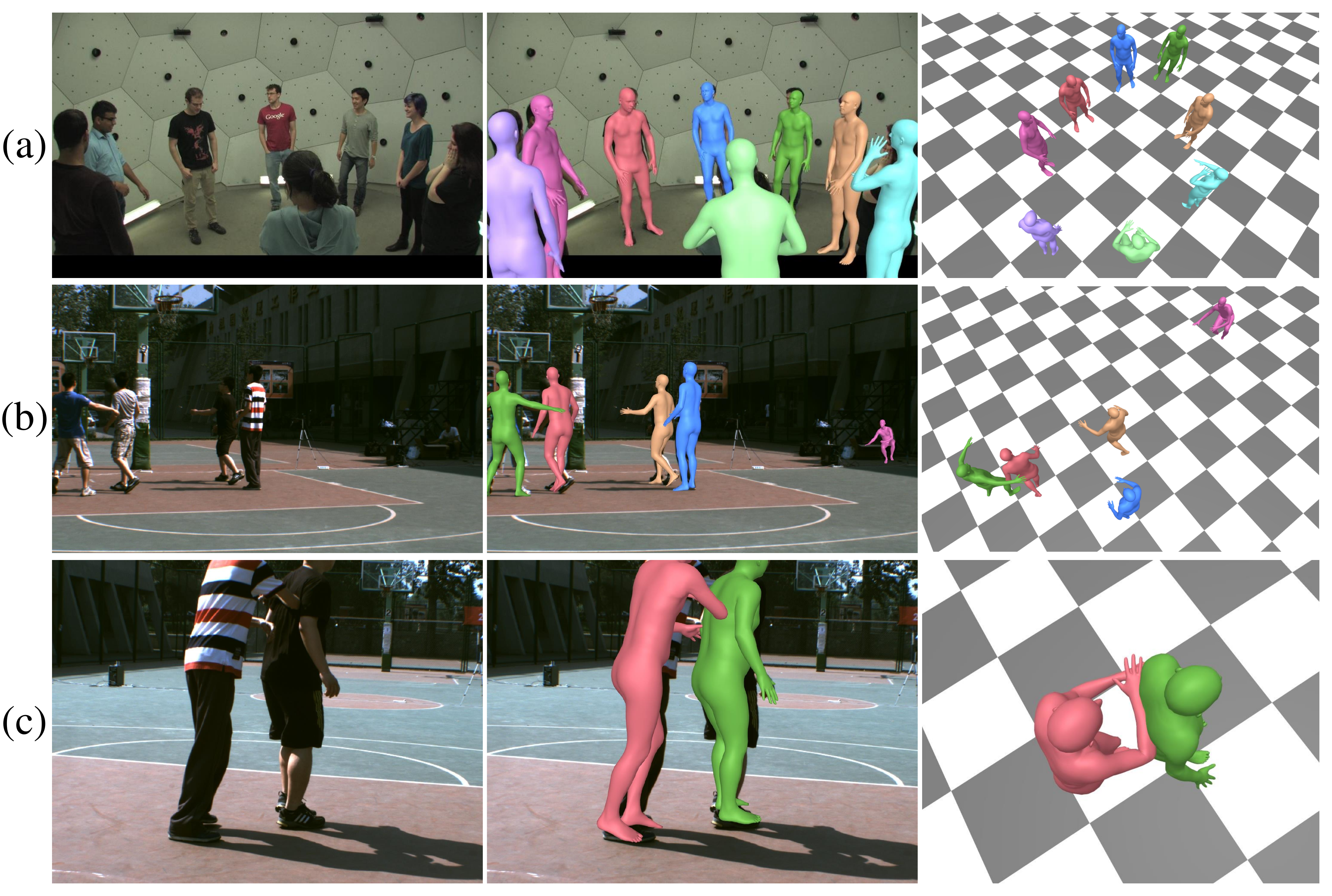}
    \end{center}
    \vspace{-4mm}
    \caption{We conducted qualitative experiments to assess the performance of our simultaneous optimization approach in scenarios with a larger number of individuals. These experiments were carried out on Panoptic~(a) and 3DMPB dataset~(b). Additionally, we evaluated our method in a close-range viewpoint setting using 3DMPB dataset~(c). The results demonstrate the adaptability of our method to more complex environments.}
    \label{fig:morepeople}
    \vspace{-3mm}
\end{figure}

\begin{figure}
    \begin{center}
    \includegraphics[width=1.0\linewidth]{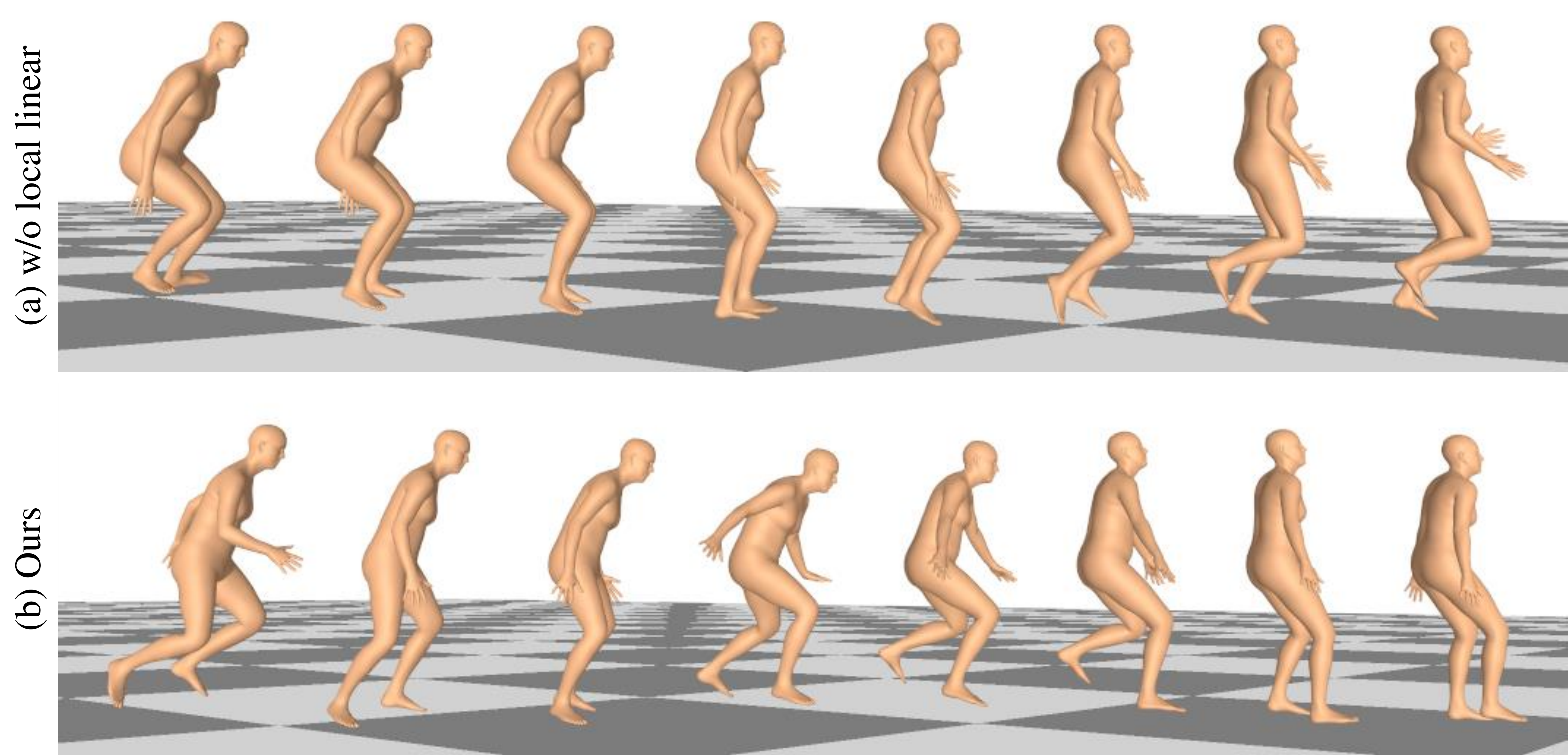}
    \end{center}
    \vspace{-5mm}
    \caption{Our motion prior can also be applied to generative tasks. With the local linear constraint, the motion prior can generate natural and plausible meshes.}
    \label{fig:generative}
\vspace{-5mm}
\end{figure}

\noindent\textbf{Motion prior}. Although our motion prior is train on short motion clips~(\ie, 16 frames), its recurrent structure enables it to effectively model long motions, such as those comprising 384 frames on the MHHI dataset. Furthermore, in contrast to explicit SMPL parameters, our motion prior employs a latent representation, reducing the number of optimization variables in motion capture by 53\%. To demonstrate its superiority, we conducted a comparison with VPoser-t, which is a combination of VPoser~\cite{pavlakos2019expressive}, as shown in \tabref{tab:ablation}. Since Vposer-t lacks global dynamics, the results show that the standard variance of our method on MHHI is smaller. \tabref{tab:camera Calibration}, \tabref{tab:ablation}, and \figref{fig:ablation} demonstrate that VPoser-t is more sensitive to noisy detections due to the lack of temporal constraints. Additionally, the local linear constraint plays a crucial role in maintaining smooth transitions between each frame of the latent code. To investigate its impact, we conducted experiments with and without the local linear constraint during motion prior training. As shown in \tabref{tab:ablation}, the absence of the local linear constraint resulted in a small gap on mean distance error on MHHI dataset. However, it led to a significantly larger standard variance. This outcome highlights the effectiveness of the constraint in modeling temporally coherent motions. In order to assess the effectiveness of other components, we further conducted ablation studies as presented in \tabref{tab:ablation_motionprior}. We first replace the bidirectional GRU with a basic backbone that contains a single GRU. The difference in mean distance error between "w/ basic GRU" and "Motion Prior" demonstrates that the bidirectional GRU is more effective in modeling long-term motion prior knowledge. Additionally, "w/o skip-connection" refers to a model in which the skip-connection in the motion prior has been removed. Without this module, the prior predominantly focuses on global motion information, potentially hindering the reconstruction of local fine-grained poses. We also conducted an experiment to demonstrate the generative ability of our motion prior. Randomly sampling the latent code for each frame will lead to an incoherent motion. Thus, we sampled the latent code of the start and end frames from the standard Gaussian distribution and generated the code of the entire motion with linear interpolation. \figref{fig:generative} compares the motion prior with and without the local linear constraint. The results show that the interpolated latent code will produce preternatural interpenetration without the constraint. On the contrary, with the local linear constraint, we can generate temporal coherent and diverse motions by linearly interpolating the sampled latent code.

\noindent\textbf{Camera estimation}. The results in \tabref{tab:camera Calibration} show that the camera parameters are significantly improved with the simultaneous reconstruction, which can achieve a similar performance to the ground-truth. We further compared the joint accuracy of the simultaneous reconstruction to the motion capture with ground-truth cameras in \tabref{tab:ablation}. With the influence of less accurate cameras, the motion capture can still achieve satisfactory performance. Besides, we also compared our simultaneous optimization with the separate estimation method. In \tabref{tab:ablation}, we observed that the performance degradation of the separate estimation method on the Shelf is greater than that on the Campus. This difference can be attributed to the presence of more complex occlusions on Shelf dataset. In contrast, the calibration and reconstruction can promote each other in our simultaneous optimization, making the framework more robust to noisy inputs.

\noindent\textbf{Robustness to occlusions}. Occlusions are common in real-world scenarios. To explicitly evaluate the performance of our method on the occlusion dataset, we conducted the experiments on OcMotion dataset, which contains real object occlusions. Although it is a multi-view dataset, we use the monocular video as input for evaluating the occlusion sensitivity. The quantitative results on multi-person datasets in \tabref{tab:ablation} demonstrate that our method is more robust to the occlusions than VPoser-t.

\noindent\textbf{Scalability}. Due to the flexibility of association with respect to the number of people and the representation of motion by independent motion priors, our method can be extended to accommodate scenes with a higher number of individuals. In \figref{fig:morepeople}~(a, b), we present qualitative results obtained from the Panoptic and 3DMPB datasets, each featuring 8 and 5 people, respectively. Our method consistently delivers satisfactory results in these scenarios. Additionally, we evaluate our method from a close-range viewpoint, as depicted in \figref{fig:morepeople}~(c). Despite significant deviations between the initial camera parameters and the actual values in such views, our simultaneous optimization still yields favorable outcomes.

%% file: conclusion.tex
\section{Conclusion}\label{sec:conclusion}
% \vspace{-2mm}
This paper proposes a framework that directly recovers human motions and camera parameters from sparse multi-view video cameras. We introduce a camera initialization based on pure human cues for simultaneous reconstruction. Unlike previous work, which fails to establish view-view and model-view correspondences with less accurate cameras, we introduce a pose-geometry consistent association to select correct 2D poses from the detected human semantics. In addition, we also propose a novel latent motion prior to jointly optimize camera parameters and coherent human motions from slightly noisy inputs. The proposed method simplifies conventional multi-person mesh recovery by incorporating calibration and reconstruction into a one-step reconstruction framework.

However, it's important to note that our method does have some limitations. First, knowledge of the exact number of people in the scene is required as we need to assign a motion prior to each individual. Second, the current method relies on human semantics for camera estimation. Consequently, the framework may encounter challenges when there is minimal or no visible portion of a person in the image. Lastly, the current implementation is limited to static cameras and does not support moving cameras. In the future, to incorporate physical priors to decouple camera motions and human dynamics for moving cameras and perform the calibration and mesh recovery in large-scale scenarios may be a prospective direction. We view these limitations as opportunities for future research, providing fertile ground for further exploration and expansion of our simultaneous optimization framework.